\newcommand\SEKIusersusepackages
{\usepackage{url}}
\newcommand\SEKIREVIEWSTATUS{1}
\newcommand\SEKIREVIEWER{\\\huttername
\\DFKI\,GmbH, Stuhlsatzenhausweg\,3, \plzuniSB\,\SB, Germany
\\E-mail: {\tt hutter@dfki.de}
\\WWW: \url{http://www.dfki.de/~hutter}
\\\\\padawitzname
\\\addressuniDOcomplete
\\E-mail: {\tt  peter.padawitz@udo.edu}
\\WWW: \url{http://fldit-www.cs.uni-dortmund.de/~peter}
\\\\\samoaname
\\\addressuniKLcomplete
\\E-mail: {\tt t.samoa@netcologne.de}
\\WWW: \url{http://agent.informatik.uni-kl.de/mitarbeiter/schmidt-samoa}
\\\mbox{}\\\mbox{}\\\mbox{}}
\newcommand\SEKIDOCUMENTCLASS{0}
\newcommand\SEKIYEAR{2006}
\newcommand\SEKINUMBEROFYEAR{01}
\newcommand\SEKITYPE{1}
\newcommand\SEKIPUBLISHEDTYPE{0}
\newcommand\daspaper{paper}
\newcommand\DasPaper{Paper}
\newcommand\encodingstext{some form of induction. \
(Such an encoding can be found, \eg, 
 in the induction rule of \nolinebreak\cite{gentzenconsistent}, 
 or by application of the \secondorder\ \theoremofnoetherianinduction\
 \nlbmath{\inpit{\ident{N}}} (\cfnlb\ \sectref{section descente infinie}),
 or the \secondorder\ \axiomofstructuralinduction\
 \nlbmath{\inpit{\ident{S}}} (\cfnlb\ \sectref{section descente infinie}),
 or by generation of \firstorder\ induction axioms.)}
\newcommand\ripplingcitation{\cite{rippling-calculus,rippling,rippling-book,hutter-rippling,hutter-diss,coloringterms,higher-order-annotated}}
\newcommand\admissibilitycitation
{\cite{quodlibet-cade,kwspec,wirth-shallow,wgcade}}
\newcommand\oysterclamcitation
{\cite{bundy-inductive-proof-planning,oysterclam}}
\newcommand\oysterclamwithcitation
{\OYSTERCLAM\ \nolinebreak\oysterclamcitation}
\newcommand\inkawithcitation{\INKA\ \nolinebreak\cite{inkafuenf}}
\newcommand\lambdaclamcitation{\cite{bundy-survey}}
\newcommand\lambdaclamwithcitation
{\LAMBDACLAM\ \nolinebreak\lambdaclamcitation}
\newcommand\omegacitation{\cite{omegacadetwo}}
\newcommand\omegawithcitation{\OMEGA\ \nolinebreak\omegacitation}
\newcommand\acltwowithcitation{\ACLTWO\ \nolinebreak\cite{ACLTWO}}
\newcommand\quodlibetwithcitation
{\QUODLIBET\ \nolinebreak\cite{quodlibet-cade}}
\newcommand\nqthmwithcitation
{\NQTHM\ \nolinebreak\cite{bm,boyermoore}}
\newcommand\Schoolofexplicitinduction{School of Explicit Induction}
\newcommand\Schoolofimplicitinduction{School of Implicit Induction}
\newcommand\thehardtasks{the two hard tasks 
mentioned in \nlbsectref{section descente infinie} \hskip.1em
(namely the Hypotheses Task and the Induction-Ordering Task)}
\title
{Progress in Computer-Assisted
\\Inductive Theorem Proving 
\\by Human-Orientedness
\\and\emph\DescenteInfinie?}
\author{\wirthname\\\Institute\\\tt\email}
\input seki-deckblatt-2

\usepackage{url}

\newlength{\mybibitemsep}
\newcommand\setmybibitemsep[1]{\setlength{\mybibitemsep}{#1}}

\setmybibitemsep{0.5\topsep}

\newcommand\includenetreferences{y}

\newcommand\referencessize{\normalsize}

\newcommand\mybibbaselinestretch{0.96}

\newcommand\mybibsection[1]{\section*{#1}\if 
\addcontentslineofbibsection\addcontentsline{toc}{section}{#1}\fi}
\newcommand\addcontentslineofbibsection{y}

\newcommand\mybibtitle[1]{{\em #1\@.}}
\newcommand\mybibhardnodate
[1]{{\def\myfirstarg{#1}\ifx\empty\myfirstarg{}\else#1.\ \fi}}
\newcommand\mybibsoft
[2]{{\def\myfirstarg{#1}\def\mysecondarg{#2}\ifx\empty\myfirstarg{}\else
\if y\includenetreferences\writemynetsource#1\else\if x\includenetreferences
\ifx\empty\mysecondarg\writemynetsource#1\else{}\fi\else
\if n\includenetreferences{}\else\typeout
{WARNING includenetreferences from headerbiblio.tex has silly definition}
\fi\fi\fi\fi}}

\def\writemynetsource[#1,#2,#3,#4://#5]{{\tt\sloppy\ 
\url{#4://#5} \discretionary
{(\ignorespaces#2\,\ignorespaces#1\mbox{$\!$},\mbox{$\!$}}%
{\ignorespaces#3)\mbox{$\!$}.}%
{(\ignorespaces#2\,\ignorespaces#1\mbox{$\!$},\,\ignorespaces
#3)\mbox{$\!$}.}}}

\catcode`\@=11

\newcommand\resetlongbibstyle{%
\def\bibitem{\@ifnextchar[\@lbibitem\@bibitem}
\def\@lbibitem[##1]##2{\item[\@biblabel{##1}\hfill]\if@filesw
      {\let\protect\noexpand
       \immediate
       \write\@auxout{\string\bibcite{##2}{##1}}}\fi\ignorespaces}
\def\@bibitem##1{\item\if@filesw \immediate\write\@auxout
       {\string\bibcite{##1}{\the\value{\@listctr}}}\fi\ignorespaces}
\def\bibcite{\@newl@bel b}
\let\citation\@gobble
\let\bibdata=\@gobble
\let\bibstyle=\@gobble
\def\bibliography##1{%
  \if@filesw
    \immediate\write\@auxout{\string\bibdata{##1}}%
  \fi
  \@input@{\jobname.bbl}}
\def\bibliographystyle##1{%
  \ifx\@begindocumenthook\@undefined\else
    \expandafter\AtBeginDocument
  \fi
    {\if@filesw
       \immediate\write\@auxout{\string\bibstyle{##1}}%
     \fi}}
\def\nocite##1{\@bsphack
  \@for\@citeb:=##1\do{%
    \edef\@citeb{\expandafter\@firstofone\@citeb}%
    \if@filesw\immediate\write\@auxout{\string\citation{\@citeb}}\fi
    \@ifundefined{b@\@citeb}{\G@refundefinedtrue
        \@latex@warning{Citation `\@citeb' undefined}}{}}%
  \@esphack}
\expandafter\let\csname b@*\endcsname\@empty
\DeclareRobustCommand\cite{%
  \@ifnextchar [{\@tempswatrue\@citex}{\@tempswafalse\@citex[]}}
\DeclareRobustCommand\citet{%
  \@ifnextchar [{\@tempswatrue\@citex}{\@tempswafalse\@citex[]}}
\def\@tempswafalse{\let\if@tempswa\iffalse}
\def\@tempswatrue{\let\if@tempswa\iftrue}
\let\if@tempswa\iffalse
\def\@cite##1##2{##1\if@tempswa , ##2\fi}
\def\@citex[##1]##2{%
  \let\@citea\@empty
  \@cite{\@for\@citeb:=##2\do
    {\@citea\def\@citea{,\penalty\@m\ }%
     \edef\@citeb{\expandafter\@firstofone\@citeb}%
     \if@filesw\immediate\write\@auxout{\string\citation{\@citeb}}\fi
     \@ifundefined{b@\@citeb}{\mbox{\reset@font\bfseries ?}%
       \G@refundefinedtrue
       \@latex@warning
         {Citation `\@citeb' on page \thepage \space undefined}}%
       {\csname b@\@citeb\endcsname}}}{##1}}
\def\@biblabel##1{##1}
\def\mybibitem##1##2##3##4##5##6##7##8##9
{\item
 \if@filesw
      {\let\protect\noexpand
       \immediate
       \write\@auxout{\string\bibcite{##1}{##7\discretionary
{}{}{\,}(##3##9)}}}\fi\ignorespaces
##2 (##3##9). \mybibtitle{##4} \mybibhardnodate{##5}\mybibsoft
{##6}{##5}$\!\!$\par}
\def\thebibliography##1{\mybibsection{\refname}%
\@mkboth{\uppercase{\refname}}{\uppercase{\refname}}
\def\baselinestretch{\mybibbaselinestretch}%
\list{}{\labelwidth\z@
    \leftmargin 1.5pc
    \itemindent-\leftmargin}
    \referencessize
    \parindent\z@
    \parskip\mybibitemsep\relax
    \def\newblock{\hskip .11em plus .33em minus .07em}
    \sloppy\clubpenalty4000\widowpenalty4000
    \sfcode`\.=1000\relax}
\let\endthebibliography=\endlist
}

\newcommand\resetshortbibstyle{%
\def\bibitem{\@ifnextchar[\@lbibitem\@bibitem}
\def\@lbibitem[##1]##2{\item[\@biblabel{##1}\hfill]\if@filesw
      {\let\protect\noexpand
       \immediate
       \write\@auxout{\string\bibcite{##2}{##1}}}\fi\ignorespaces}
\def\@bibitem##1{\item\if@filesw \immediate\write\@auxout
       {\string\bibcite{##1}{\the\value{\@listctr}}}\fi\ignorespaces}
\def\bibcite{\@newl@bel b}
\let\citation\@gobble
\def\@citex[##1]##2{%
  \let\@citea\@empty
  \@cite{\@for\@citeb:=##2\do
    {\@citea\def\@citea{,\penalty\@m\ }%
     \edef\@citeb{\expandafter\@firstofone\@citeb}%
     \if@filesw\immediate\write\@auxout{\string\citation{\@citeb}}\fi
     \@ifundefined{b@\@citeb}{\mbox{\reset@font\bfseries ?}%
       \G@refundefinedtrue
       \@latex@warning
         {Citation `\@citeb' on page \thepage \space undefined}}%
       {\hbox{\csname b@\@citeb\endcsname}}}}{##1}}
\let\bibdata=\@gobble
\let\bibstyle=\@gobble
\def\bibliography##1{%
  \if@filesw
    \immediate\write\@auxout{\string\bibdata{##1}}%
  \fi
  \@input@{\jobname.bbl}}
\def\bibliographystyle##1{%
  \ifx\@begindocumenthook\@undefined\else
    \expandafter\AtBeginDocument
  \fi
    {\if@filesw
       \immediate\write\@auxout{\string\bibstyle{##1}}%
     \fi}}
\def\nocite##1{\@bsphack
  \@for\@citeb:=##1\do{%
    \edef\@citeb{\expandafter\@firstofone\@citeb}%
    \if@filesw\immediate\write\@auxout{\string\citation{\@citeb}}\fi
    \@ifundefined{b@\@citeb}{\G@refundefinedtrue
        \@latex@warning{Citation `\@citeb' undefined}}{}}%
  \@esphack}
\expandafter\let\csname b@*\endcsname\@empty
\def\@cite##1##2{[{##1\if@tempswa , ##2\fi}]}
\def\@biblabel##1{[##1]}
\DeclareRobustCommand\cite{%
  \@ifnextchar [{\@tempswatrue\@citex}{\@tempswafalse\@citex[]}}
\def\@tempswafalse{\let\if@tempswa\iffalse}
\def\@tempswatrue{\let\if@tempswa\iftrue}
\let\if@tempswa\iffalse
\def\mybibitem##1##2##3##4##5##6##7##8##9
{\bibitem[##8##9]{##1}##2 (##3). 
\mybibtitle{##4} \mybibhardnodate{##5}\mybibsoft{##6}{##5}$\!\!$\par}
\def\thebibliography##1{\mybibsection{\refname}%
\@mkboth{\uppercase{\refname}}{\uppercase{\refname}}%
\def\baselinestretch{\mybibbaselinestretch}%
\referencessize
\list
 {[\arabic{enumi}]}
 {\settowidth\labelwidth{[##1]}\leftmargin\labelwidth
 \advance\leftmargin\labelsep
 \usecounter{enumi}}
 \parskip\mybibitemsep\relax
 \def\newblock{\hskip .11em plus .33em minus .07em}
 \sloppy\clubpenalty4000\widowpenalty4000
 \sfcode`\.=1000\relax}
\let\endthebibliography=\endlist
}

\newcommand\resetnumberbibstyle{%
\def\bibitem{\@ifnextchar[\@lbibitem\@bibitem}
\def\@lbibitem[##1]##2{\item[\@biblabel{##1}\hfill]\if@filesw
      {\let\protect\noexpand
       \immediate
       \write\@auxout{\string\bibcite{##2}{##1}}}\fi\ignorespaces}
\def\@bibitem##1{\item\if@filesw \immediate\write\@auxout
       {\string\bibcite{##1}{\the\value{\@listctr}}}\fi\ignorespaces}
\def\bibcite{\@newl@bel b}
\let\citation\@gobble
\def\@citex[##1]##2{%
  \let\@citea\@empty
  \@cite{\@for\@citeb:=##2\do
    {\@citea\def\@citea{,\penalty\@m\ }%
     \edef\@citeb{\expandafter\@firstofone\@citeb}%
     \if@filesw\immediate\write\@auxout{\string\citation{\@citeb}}\fi
     \@ifundefined{b@\@citeb}{\mbox{\reset@font\bfseries ?}%
       \G@refundefinedtrue
       \@latex@warning
         {Citation `\@citeb' on page \thepage \space undefined}}%
       {\hbox{\csname b@\@citeb\endcsname}}}}{##1}}
\let\bibdata=\@gobble
\let\bibstyle=\@gobble
\def\bibliography##1{%
  \if@filesw
    \immediate\write\@auxout{\string\bibdata{##1}}%
  \fi
  \@input@{\jobname.bbl}}
\def\bibliographystyle##1{%
  \ifx\@begindocumenthook\@undefined\else
    \expandafter\AtBeginDocument
  \fi
    {\if@filesw
       \immediate\write\@auxout{\string\bibstyle{##1}}%
     \fi}}
\def\nocite##1{\@bsphack
  \@for\@citeb:=##1\do{%
    \edef\@citeb{\expandafter\@firstofone\@citeb}%
    \if@filesw\immediate\write\@auxout{\string\citation{\@citeb}}\fi
    \@ifundefined{b@\@citeb}{\G@refundefinedtrue
        \@latex@warning{Citation `\@citeb' undefined}}{}}%
  \@esphack}
\expandafter\let\csname b@*\endcsname\@empty
\def\@cite##1##2{[{##1\if@tempswa , ##2\fi}]}
\def\@biblabel##1{[##1]}
\DeclareRobustCommand\cite{%
  \@ifnextchar [{\@tempswatrue\@citex}{\@tempswafalse\@citex[]}}
\def\@tempswafalse{\let\if@tempswa\iffalse}
\def\@tempswatrue{\let\if@tempswa\iftrue}
\let\if@tempswa\iffalse
\def\mybibitem##1##2##3##4##5##6##7##8##9
{\bibitem{##1}##2 (##3). 
\mybibtitle{##4} \mybibhardnodate{##5}\mybibsoft{##6}{##5}$\!\!$\par}
\def\thebibliography##1{\mybibsection{\refname}%
\@mkboth{\uppercase{\refname}}{\uppercase{\refname}}
\def\baselinestretch{\mybibbaselinestretch}%
\referencessize
\list
 {[\arabic{enumi}]}
 {\settowidth\labelwidth{[##1]}\leftmargin\labelwidth
 \advance\leftmargin\labelsep
 \usecounter{enumi}}
 \parskip\mybibitemsep\relax
 \def\newblock{\hskip .11em plus .33em minus .07em}
 \sloppy\clubpenalty4000\widowpenalty4000
 \sfcode`\.=1000\relax}
\let\endthebibliography=\endlist
}

\newcommand\setnumberbibstyle{\AtBeginDocument{\resetnumberbibstyle}}

\catcode`\@=12

\input headerhot

\newcommand\Proofof{Proof of}
\renewenvironment{proof}[1]{\yestop\begin{sloppypar}\noindent
{\bf\Proofof\ {#1}}\\}{\end{sloppypar}}
\renewenvironment{proofqed}[1]
{\begin{sloppypar}\def\fooqed{#1}\noindent{\bf\Proofof\ \fooqed}}
{\QEDbf\fooqed\end{sloppypar}}
\renewenvironment{proofparsep}[1]{\parindent=0pt\begin
{sloppypar}\noindent{\bf\Proofof\ {#1}}\nopagebreak\par}
{\end{sloppypar}}
\renewenvironment{proofparsepqed}[1]{\parindent=0pt\begin
{sloppypar}\def\fooqed{#1}\noindent{\bf\Proofof\ \fooqed}\nopagebreak\par}
{\nopagebreak\QEDbf\fooqed\end{sloppypar}}
\renewenvironment{proofshort}[1]{\begin{sloppypar}\noindent
{\bf\Proofof\ {#1}:}}{\end{sloppypar}}

{\end{enumerate}\renewcommand{\theenumi}{\arabic{enumi}}}

\mathcommand\myfootnotemark[1]{^{#1}}

\newcommand\repname{{\rm set}}
\mathapplycommand\rep\repname
\mathcommand\repr[1]{{\repname[{#1}]}}
\mathcommand\msa{\langle}
\mathcommand\mse{\rangle}
\mathcommand\msu{\,\sqcup\,}
\mathcommand\msin{{\rm\;in\;}}
\mathcommand\mssetminus{\setminus\!\!\setminus}
\mathcommand\tightmssubseteq{\sqsubseteq}
\mathcommand\mssubseteq{\ \tightmssubseteq\ }
\mathcommand\approxapprox{\approx\:\!\!\approx}
\mathcommand\quasilquasil{\,\lesssim\!\lesssim\,}
\mathcommand\quasibquasib{\,\gtrsim\!\gtrsim\,}
\mathcommand\fmul[1]{{\rm FMul}(#1)}
\mathcommand\smul[1]{{\rm SMul}(#1)}
\mathcommand\multisetwith [2]{\msa\ {#1}\ |\ {#2}\ \mse}
\mathcommand\multisetwithq[3]{\msa\ {#2}\ |_{#1}\ {#3}\ \mse}
\newcommand\superseteq     {\supseteq}

\mathcommand\quasilhd
{\mathop{\mbox{\raisebox{0.31ex}{\math\lhd}\hspace{-0.75em}\raisebox
{-0.6ex}{\math\sim}}}}
\newcommand\quasirhd{\mbox{\raisebox{0.31ex}{$\rhd$}\hspace{-0.75em}\raisebox{-0.6ex}{$\sim$}}}

\mathcommand\rhdrhd{\rhd$\hspace{-0.35em}$\rhd}
\mathcommand\lhdlhd{\lhd$\hspace{-0.21em}$\lhd}

\mathcommand\quasilhdquasilhd{\quasilhd$\hspace{-0.13em}$\quasilhd}

\newcommand\hiddensubSS{_{_{\rm SS}}}
\mathcommand\antisubsum     {\rhd\hiddensubSS}
\mathcommand\notantisubsum  {\ntriangleright\hiddensubSS}
\mathcommand\subsum         {\lhd\hiddensubSS}
\mathcommand\notsubsum      {\ntriangleleft\hiddensubSS}
\mathcommand\antisubsumeq   {\trianglerighteq\hiddensubSS}
\mathcommand\subsumeq       {\trianglelefteq\hiddensubSS}
\mathcommand\quasisubsum    {\,\quasilhd\raisebox{0.1ex}{$\hiddensubSS$}}
\mathcommand\antiquasisubsum{\,\quasirhd\raisebox{0.1ex}{$\hiddensubSS$}}
\mathcommand\quasiquasisubsum{\quasisubsum\!\!\quasisubsum}
\mathcommand\antiquasiquasisubsum{\antiquasisubsum\!\!\!\antiquasisubsum}

\newcommand\hiddensubH{_{_{\rm H}}}
\newcommand\hiddensubCONS{_{_\CONS}}
\mathcommand\hql   {\,\lesssim\hiddensubH}
\mathcommand\consql{\,\lesssim\hiddensubCONS}
\mathcommand\hl    {\,<       \hiddensubH}
\mathcommand\hleq  {\,\leq    \hiddensubH}
\mathcommand\consl {\,<       \hiddensubCONS}
\mathcommand\conseq{\,\approx \hiddensubCONS}

\newcommand\cons {{\rm cons}}
\mathcommand\sigconsV{\sig/\cons/\V}
\mathcommand\sigconsR{\sig/\cons/\R}
\mathcommand\primesigconsV{\sig'\!/\cons'\!/\V'}
\mathcommand\primesigconsR{\sig'\!/\cons'\!/\R'}
\newcommand\dunno{{\rm sc}}
\newcommand\DUNNO{{\rm SC}}
\mathcommand\SIGCONS   {\{\SIG,\CONS\}}
\mathcommand\sigsortstimes{\SIGCONS\tight\times\sigsorts}

\mathcommand\TERMSSYM
{\mathcal{T\hspace{-.20em}E\hspace{-.08em}R\hspace{-.16em}M\hspace{-.10em}S}}
\mathapplycommand\condterms{\TERMSSYM}
\mathcommand\kurzregel{((l,r),C)}
\mathcommand\kurzregelprime{((l',r'),C')}
\mathcommand\kurzregelindex[1]{((l_{#1},r_{#1}),C_{#1})}
\mathapplycommand\lhs{\rm lhs}

\mathcommand\red{\redsimple} 

\mathcommand\lemms{L}
\mathcommand\hypos{H}
\mathcommand\goals{G}
\mathcommand\lemmsprime{\lemms'}
\mathcommand\hyposprime{\hypos'}
\mathcommand\goalsprime{\goals'}
\mathcommand\lemmsprimeprime{\lemms''}
\mathcommand\hyposprimeprime{\hypos''}
\mathcommand\goalsprimeprime{\goals''}
\mathcommand\oldtriple            {(\lemms   ,\hypos   ,\goals  )}
\mathcommand\inittriple        {(\emptyset,\emptyset,\goals  )}
\mathcommand\triplehelp[1]     {(\lemms#1,\hypos#1 ,\goals#1)}
\mathcommand\tripleprime       {\triplehelp'}
\mathcommand\triplenogoalsprime{(\lemmsprime,\hyposprime,\emptyset  )}
\mathcommand\tripleprimeprime  {\triplehelp{''}}
\mathcommand\tripleindex[1]    {\triplehelp{_{#1}}}

\mathcommand\constcong[1]{\,\,\sim_{\!_{#1}}\,}

\mathapplycommand\avail{\rm\Av ail}

\def\emph#1{\/ {\itshape#1}\/}
\newcommand\tightemph[1]{\/{\itshape#1}\/}
\newcommand\openquoteemph[1]{\ ``\hskip-0.15em{\itshape#1}\/}

\newcommand\germantextsix{\fraknomath
Eng ist die Welt, und da\es\ Gehirn ist weit.
\\Leicht beieinander wohnen die Gedanken, 
\\Doch hart im Raume sto\sz en sich die Sachen.
}

\mathcommand\ident[1]{\mathsf{#1}}
\newcommand\plussymbol  {\ident{+}}
\newcommand\minussymbol {\ident{-}}
\newcommand\dividesymbol{\ident{/}}
\newcommand\timessymbol {\ident{*}}

\newcommand\nat     {\ident{nat}}

\newcommand\lists   {\ident{list}}
\newcommand\set     {\ident{set}}

\newcommand\naturalssymbol{\ident{naturals}}
\newcommand\gensymsymbol{\ident{gensym}}
\mathcommand\mbpsymbol{\ident{m\hspace{-0.055em}b\hspace{-0.045em}p}}

\newcommand\csymbol     {\ident c}
\newcommand\esymbol     {\ident e}
\newcommand\fsymbol     {\ident f}
\newcommand\gsymbol     {\ident g}
\newcommand\hsymbol     {\ident h}
\newcommand\ksymbol     {\ident k}
\newcommand\ssymbol     {\ident s}
\newcommand\Everysymbol {\ident{Every}}
\newcommand\Permsymbol {\ident{Perm}}
\newcommand\RExistssymbol{\ident{Rexists}}
\newcommand\invertsymbol{\ident{invert}}
\newcommand\invsymbol{\ident{inv}}
\newcommand\abssymbol   {\ident{abs}}
\newcommand\cnssymbol   {\ident{cons}}
\mathcommand\cnsindexsymbol[1]{\ident{cons}_{#1}}

\newcommand\lengthsymbol{\ident{length}}
\newcommand\dlsymbol    {\ident{dl}}
\newcommand\dloncesymbol{\ident{delonce}}
\newcommand\rcsymbol    {\ident{rc}}
\newcommand\brsymbol    {\ident{br}}
\newcommand\revtailsymbol{\ident{revtail}}
\newcommand\revsymbol{\ident{rev}}
\newcommand\appendsymbol {\ident{append}}
\newcommand\zeropredicatesymbol{\ident{zerop}}
\newcommand\eqsymbol        {\ident{eq}}
\newcommand\ifthensymbol    {\mbox{\ident{If{}Then}}}
\newcommand\ifthenelsesymbol{\mbox{\ident{If{}ThenElse}}}
\mathcommand\eqindexsymbol        [1]{\eqsymbol        _{#1}}
\mathcommand\ifthenindexsymbol    [1]{\ifthensymbol    _{#1}}
\mathcommand\ifthenelseindexsymbol[1]{\ifthenelsesymbol_{#1}}
\newcommand\orsymbol    {\ident{or}}
\newcommand\andsymbol   {\ident{and}}
\newcommand\leqsymbol   {\ident{leq}}
\newcommand\lessymbol   {\ident{less}}
\newcommand\lexsymbol   {\ident{lex}}
\newcommand\acksymbol   {\ident{ack}}
\newcommand\switchsymbol{\ident{switch}}
\newcommand\swatchsymbol{\ident{swatch}}
\newcommand\diveinssymbol{\ident{div1}}
\newcommand\divzweisymbol{\ident{div2}}
\newcommand\divrestsymbol{\ident{divrest}}
\newcommand\diveinstailsymbol{\ident{div1tail}}
\newcommand\divzweitailsymbol{\ident{div2tail}}

\newcommand\turingmachinesymbol{\ident T}
\newcommand\terminatespsymbol  {\ident{terminatesp}}
\newcommand\statesymbol        {\ident{state}}
\newcommand\cmdsymbol          {\ident{cmd}}
\newcommand\nthsymbol          {\ident{nth}}
\newcommand\doublesymbol       {\ident{double}}

\newcommand\ppsymbol           {\ident{p}}
\newcommand\qpsymbol           {\ident{q}}
\newcommand\Epsymbol           {\ident{E}}
\newcommand\Ppsymbol           {\ident{P}}
\newcommand\Qpsymbol           {\ident{Q}}
\newcommand\Marriessymbol      {\ident{Marries}}
\newcommand\Lovessymbol        {\ident{Loves}}
\newcommand\StolenBysymbol     {\ident{StolenBy}}
\newcommand\Humansymbol        {\ident{Human}}
\newcommand\Evensymbol         {\ident{Even}}
\newcommand\Oddsymbol          {\ident{Odd}}
\newcommand\Primesymbol        {\ident{Prime}}
\newcommand\EveryPairsymbol   {\ident{EveryPair}}
\newcommand\Givesymbol         {\ident{Give}}
\newcommand\Fathersymbol       {\ident{Father}}
\newcommand\Elephantpsymbol    {\ident{Elephant}}
\newcommand\Flowerpsymbol    {\ident{Flower}}
\newcommand\Germanpsymbol      {\ident{German}}
\newcommand\Bicyclepsymbol     {\ident{Bicycle}}
\newcommand\Hugepsymbol        {\ident{Huge}}
\newcommand\Animalpsymbol      {\ident{Animal}}
\newcommand\Malepsymbol        {\ident{Male}}
\newcommand\Boypsymbol        {\ident{Boy}}
\newcommand\Girlpsymbol        {\ident{Girl}}
\newcommand\Femalepsymbol      {\ident{Female}}
\newcommand\Roundpsymbol       {\ident{Round}}
\newcommand\Quadrangularpsymbol{\ident{Quadrangular}}
\newcommand\Metpsymbol         {\ident{Met}}
\newcommand\Bishopsymbol       {\ident{Bishop}}
\newcommand\mindexsymbol[1]{\existsvari w{#1}}

\newcommand\nonnegpsymbol      {\ident{nonnegp}}
\newcommand\wellsymbol         {\ident{well}}
\newcommand\welltailsymbol     {\ident{welltail}}
\newcommand\varsymbol          {\ident{var}}
\newcommand\aritysymbol        {\ident{arity}}

\newcommand\whilesymbol        {\ident{while}}

\newcommand\nullsymbol         {\ident{null}}
\newcommand\hdsymbol           {\ident{hd}}
\newcommand\tlsymbol           {\ident{tl}}
\newcommand\insymbol           {\ident{in}}
\newcommand\applysymbol        {\ident{app}}
\newcommand\termsymbol         {\ident{term}}
\mathcommand\tightim{\longrightarrow}
\mathcommand\im{\ \tightim\ }
\mathcommand\rs{\:\rulesugar\:\:}
\mathcommand\rulesugar{\longleftarrow}

\mathcommand\doublepp[1]      {\doublesymbol   \beginargs{#1}\allargs}
\mathcommand\aritypp[1]      {\aritysymbol   \beginargs{#1}\allargs}
\mathcommand\lengthpp[1]      {\lengthsymbol   \beginargs{#1}\allargs}
\mathcommand\wellpp[1]      {\wellsymbol   \beginargs{#1}\allargs}
\mathcommand\welltailpp[1]      {\welltailsymbol   \beginargs{#1}\allargs}
\mathcommand\varpp[1]      {\varsymbol   \beginargs{#1}\allargs}
\mathcommand\divrestpp[2]    {\divrestsymbol\beginargs{#1}\separgs{#2}\allargs}
\mathcommand\diveinspp[2]    {\diveinssymbol\beginargs{#1}\separgs{#2}\allargs}
\mathcommand\divzweipp[3]    {\divzweisymbol\beginargs{#1}\separgs{#2}
\separgs{#3}\allargs}
\mathcommand\diveinstailpp[4]    {\diveinstailsymbol\beginargs{#1}\separgs{#2}
\separgs{#3}\separgs{#4}\allargs}
\mathcommand\divzweitailpp[6]    {\divzweitailsymbol\beginargs{#1}\separgs{#2}
\separgs{#3}\separgs{#4}\separgs{#5}\separgs{#6}\allargs}
\mathcommand\mbppp[2]         {\mbpsymbol   \beginargs{#1}\separgs{#2}\allargs}
\mathcommand\revpp[1]     
{\revsymbol\beginargs{#1}\allargs}
\mathcommand\revppi[2]     
{\revsymbol^{#1}\beginargs{#2}\allargs}
\mathcommand\revtailpp[2]     
{\revtailsymbol\beginargs{#1}\separgs{#2}\allargs}
\mathcommand\revtailppi[3]
{\revtailsymbol^{#1}\beginargs{#2}\separgs{#3}\allargs}
\mathcommand\Permpp[2]     
{\Permsymbol\beginargs{#1}\separgs{#2}\allargs}
\mathcommand\Permppi[3]
{\Permsymbol^{#1}\beginargs{#2}\separgs{#3}\allargs}
\mathcommand\appendpp[2]      
{\appendsymbol \beginargs{#1}\separgs{#2}\allargs}
\mathcommand\appendppi[3]      
{\appendsymbol^{#1}\beginargs{#2}\separgs{#3}\allargs}
\mathcommand\Everypp[2]      
{\Everysymbol \beginargs{#1}\separgs{#2}\allargs}
\mathcommand\RExistspp[1]      
{\RExistssymbol \beginargs{#1}\allargs}
\mathcommand\appendlongpp[2]      
{\appendsymbol\left(\begin{array}{@{}l@{}}{#1}\separgs\\{#2}\end{array}\right)}
\mathcommand\cnspp[2]         {\cnssymbol   \beginargs{#1}\separgs{#2}\allargs}
\mathcommand\cnsppi[3]       {\cnssymbol^{#1}\beginargs{#2}\separgs{#3}\allargs}
\mathcommand\cnsindexpp[3]
{\cnsindexsymbol{#1}\beginargs{#2}\separgs{#3}\allargs}
\mathcommand\dlpp[2]          {\dlsymbol    \beginargs{#1}\separgs{#2}\allargs}
\mathcommand\dloncepp[2]      {\dloncesymbol\beginargs{#1}\separgs{#2}\allargs}
\mathcommand\dlonceppi[3]{\dloncesymbol^{#1}\beginargs{#2}\separgs{#3}\allargs}
\mathcommand\rcpp[2]          {\rcsymbol    \beginargs{#1}\separgs{#2}\allargs}
\mathcommand\brpp[2]          {\brsymbol    \beginargs{#1}\separgs{#2}\allargs}
\mathcommand\orpp[2]          {\orsymbol    \beginargs{#1}\separgs{#2}\allargs}
\mathcommand\andpp[2]         {\andsymbol   \beginargs{#1}\separgs{#2}\allargs}
\mathcommand\shortcnspp[2]    {\csymbol     \beginargs{#1}\separgs{#2}\allargs}
\mathcommand\tightshortcnspp[2]
{\csymbol\beginargs{#1}\tightsepargs{#2}\allargs}
\mathcommand\spp[1]           {\ssymbol     \beginargs{#1}\allargs}
\mathcommand\sppiterated[2]   {\ssymbol^{#1}\beginargs{#2}\allargs}
\mathcommand\ppp[1]           {\psymbol     \beginargs{#1}\allargs}
\mathcommand\pppiterated[2]   {\psymbol^{#1}\beginargs{#2}\allargs}
\mathcommand\zeropp           {\ident 0}
\mathcommand\Julietpp         {\ident{Juliet}}
\mathcommand\Romeopp          {\ident{Romeo}}
\mathcommand\Ipp              {\ident I}
\mathcommand\onepp            {\ident1}
\mathcommand\twopp            {\ident2}
\mathcommand\threepp          {\ident3}
\mathcommand\invertpp[1]      {\invertsymbol\beginargs{#1}\allargs}
\mathcommand\invpp[1]         {\invsymbol\beginargs{#1}\allargs}
\mathcommand\abspp[1]         {\abssymbol\beginargs{#1}\allargs}
\mathcommand\naturalspp[1]    {\naturalssymbol\beginargs{#1}\allargs}
\mathcommand\gensympp[1]      {\gensymsymbol\beginargs{#1}\allargs}
\mathcommand\nilpp            {\ident{nil}}
\mathcommand\falsepp          {\ident{false}}
\mathcommand\truepp           {\ident{true}}
\mathcommand\FALSEpp          {\ident{FALSE}}
\mathcommand\TRUEpp           {\ident{TRUE}}
\mathcommand\weirdppp         {\ident{weirdp}}
\mathcommand\ambigppp         {\ident{ambigp}}
\mathcommand\zeropredicatepp[1]{\zeropredicatesymbol\beginargs{#1}\allargs}
\mathcommand\cppeins       [1]{\csymbol     \beginargs{#1}\allargs}
\mathcommand\cppzwei       [2]{\csymbol\beginargs{#1}\separgs{#2}\allargs}
\mathcommand\eppeins       [1]{\esymbol     \beginargs{#1}\allargs}
\mathcommand\fppeins       [1]{\fsymbol     \beginargs{#1}\allargs}
\mathcommand\fppeinsindex  [2]{\fsymbol_{#1}\beginargs{#2}\allargs}
\mathcommand\fppeinsiterated[2]{\fsymbol^{#1}\beginargs{#2}\allargs}
\mathcommand\gppeins       [1]{\gsymbol     \beginargs{#1}\allargs}
\mathcommand\gppzwei       [2]{\gsymbol     \beginargs{#1}\separgs{#2}\allargs}
\mathcommand\hppeins       [1]{\hsymbol     \beginargs{#1}\allargs}
\mathcommand\kppeins       [1]{\ksymbol     \beginargs{#1}\allargs}
\mathcommand\appzero          {\ident a}
\mathcommand\bppzero          {\ident b}
\mathcommand\cppzero          {\ident c}
\mathcommand\dppzero          {\ident d}
\mathcommand\eppzero          {\ident e}
\mathcommand\eqindexpp[3]{\eqindexsymbol{#1}\beginargs{#2}\separgs{#3}\allargs}
\mathcommand\ifthenindexpp
[3]{\ifthenindexsymbol{#1}\beginargs{#2}\separgs{#3}\allargs}
\mathcommand\ifthenelseindexpp
[4]{\ifthenelseindexsymbol{#1}\beginargs{#2}\separgs{#3}\separgs{#4}\allargs}
\mathcommand\eqpp[2]{\eqsymbol\beginargs{#1}\separgs{#2}\allargs}
\mathcommand\leqpp[2]{\leqsymbol\beginargs{#1}\separgs{#2}\allargs}
\mathcommand\lespp[2]{\lessymbol\beginargs{#1}\separgs{#2}\allargs}
\mathcommand\lexpp[3]{\lexsymbol\beginargs{#1}\separgs{#2}\separgs{#3}\allargs}
\mathcommand\ackpp[2]{\acksymbol\beginargs{#1}\separgs{#2}\allargs}
\mathcommand\switchpp[1]{\switchsymbol\beginargs{#1}\allargs}
\mathcommand\swatchpp[1]{\swatchsymbol\beginargs{#1}\allargs}
\mathcommand\whilepp[2]{\whilesymbol\beginargs{#1}\separgs{#2}\allargs}
\mathcommand\nullpp[1]{\nullsymbol\beginargs{#1}\allargs}
\mathcommand\nullppiterated[2]{\nullsymbol^{#1}\beginargs{#2}\allargs}
\mathcommand\hdpp[1]{\hdsymbol\beginargs{#1}\allargs}
\mathcommand\hdppiterated[2]{\hdsymbol^{#1}\beginargs{#2}\allargs}
\mathcommand\tlpp[1]{\tlsymbol\beginargs{#1}\allargs}
\mathcommand\tlppiterated[2]{\tlsymbol^{#1}\beginargs{#2}\allargs}
\mathcommand\inpp[2]{\insymbol\beginargs{#1}\separgs{#2}\allargs}
\mathcommand\inppiterated[3]{\insymbol^{#1}\beginargs{#2}\separgs{#3}\allargs}
\mathcommand\applypp[2]{\applysymbol\beginargs{#1}\separgs{#2}\allargs}
\mathcommand\applyppiterated
[3]{\applysymbol^{#1}\beginargs{#2}\separgs{#3}\allargs}
\mathcommand\termpp[2]{\termsymbol\beginargs{#1}\separgs{#2}\allargs}
\mathcommand\setpp[1]{\set\beginargs{#1}\allargs}

\mathcommand\Tpp[6]{\turingmachinesymbol\beginargs{#1}\separgs{#2}\separgs
{#3}\separgs{#4}\separgs{#5}\separgs{#6}\allargs}
\mathcommand\Tppseven[7]{\turingmachinesymbol\beginargs{#1}\separgs{#2}\separgs
{#3}\separgs{#4}\separgs{#5}\separgs{#6}\separgs{#7}\allargs}
\mathcommand\foreverppp[6]{\ident{foreverp}\beginargs{#1}\separgs{#2}\separgs
{#3}\separgs{#4}\separgs{#5}\separgs{#6}\allargs}
\mathcommand\terminatesppp[6]{\terminatespsymbol\beginargs{#1}\separgs
{#2}\separgs{#3}\separgs{#4}\separgs{#5}\separgs{#6}\allargs}
\mathcommand\terminatespppone[1]{\terminatespsymbol \beginargs{#1}\allargs}
\mathcommand\statepp
[3]{\statesymbol\beginargs{#1}\separgs{#2}\separgs{#3}\allargs}
\mathcommand\tightstatepp
[3]{\statesymbol\beginargs{#1}\tightsepargs{#2}\tightsepargs{#3}\allargs}
\mathcommand\cmdpp
[3]{\cmdsymbol  \beginargs{#1}\separgs{#2}\separgs{#3}\allargs}
\mathcommand\tightcmdpp
[3]{\cmdsymbol  \beginargs{#1}\tightsepargs{#2}\tightsepargs{#3}\allargs}
\mathcommand\stoppp           {\ident{stop}}
\mathcommand\leftpp           {\ident{left}}
\mathcommand\rightpp          {\ident{right}}
\mathcommand\nthpp         [2]{\nthsymbol  \beginargs{#1}\separgs{#2}\allargs}
\mathcommand\pppp          [1]{\ppsymbol\beginargs{#1}            \allargs}
\mathcommand\qppp          [2]{\qpsymbol\beginargs{#1}\separgs{#2}\allargs}
\mathcommand\Eppp          [1]{\Epsymbol\beginargs{#1}            \allargs}
\mathcommand\Epppzwei      [2]{\Epsymbol\beginargs{#1}\separgs{#2}\allargs}
\mathcommand\Pppp          [1]{\Ppsymbol\beginargs{#1}            \allargs}
\mathcommand\Ppppdrei      
[3]{\Ppsymbol\beginargs{#1}\separgs{#2}\separgs{#3}\allargs}
\mathcommand\Ppppvier
[4]{\Ppsymbol\beginargs{#1}\separgs{#2}\separgs{#3}\separgs{#4}\allargs}
\mathcommand\Qppp          [2]{\Qpsymbol\beginargs{#1}\separgs{#2}\allargs}
\mathcommand\Qpppeins      [1]{\Qpsymbol\beginargs{#1}\allargs}
\mathcommand\Qpppdrei      
[3]{\Qpsymbol\beginargs{#1}\separgs{#2}\separgs{#3}\allargs}
\mathcommand\Fatherpp      [2]{\Fathersymbol\beginargs{#1}\separgs{#2}\allargs}
\mathcommand\Marriespp     [2]{\Marriessymbol\beginargs{#1}\separgs{#2}\allargs}
\mathcommand\Lovespp       [2]{\Lovessymbol\beginargs{#1}\separgs{#2}\allargs}
\mathcommand\StolenBypp    [2]
{\StolenBysymbol\beginargs{#1}\separgs{#2}\allargs}
\mathcommand\Humanpp       [1]{\Humansymbol\beginargs{#1}\allargs}
\mathcommand\Evenpp        [1]{\Evensymbol\beginargs{#1}\allargs}
\mathcommand\Evenppi       [2]{\Evensymbol^{#1}\beginargs{#2}\allargs}
\mathcommand\Oddpp         [1]{\Oddsymbol\beginargs{#1}\allargs}
\mathcommand\Primepp       [1]{\Primesymbol\beginargs{#1}\allargs}
\mathcommand\EveryPairpp  [2]{\EveryPairsymbol\beginargs{#1}\separgs
{#2}\allargs}
\mathcommand\mindexppeins  [2]{\mindexsymbol{#1}\beginargs{#2}\allargs}
\mathcommand\Givepp        [3]{\Givesymbol
\beginargs{#1}\separgs{#2}\separgs{#3}\allargs}
\mathcommand\mindexppzwei  [3]{\mindexsymbol
{#1}\beginargs{#2}\separgs{#3}\allargs}
\mathcommand\mindexppdrei  [4]{\mindexsymbol
{#1}\beginargs{#2}\separgs{#3}\separgs{#4}\allargs}

\mathcommand\nonnegppp     [1]{\nonnegpsymbol\beginargs{#1}\allargs}

\mathcommand\anonymouscsymbol{c}
\mathcommand\anonymouscindexsymbol[1]{\anonymouscsymbol_{#1}}
\mathcommand\anonymousfsymbol{f}
\mathcommand\anonymouscpp
[2]{\anonymouscsymbol\beginargs{#1}\separgs\ldots\separgs{#2}\allargs}
\mathcommand\anonymouscindexpp
[3]{\anonymouscindexsymbol{#1}\beginargs{#2}\separgs\ldots\separgs{#3}\allargs}
\mathcommand\anonymousfpp
[2]{\anonymousfsymbol\beginargs{#1}\separgs\ldots\separgs{#2}\allargs}
\mathcommand\coerceindexpp[3]{[#3]_{#1}^{#2}}

\mathcommand\Elephantppp    [1]{\Elephantpsymbol\beginargs{#1}\allargs}
\mathcommand\Flowerppp      [1]{\Flowerpsymbol  \beginargs{#1}\allargs}
\mathcommand\Bicycleppp     [1]{\Bicyclepsymbol \beginargs{#1}\allargs}
\mathcommand\Germanppp      [1]{\Germanpsymbol  \beginargs{#1}\allargs}
\mathcommand\Hugeppp        [1]{\Hugepsymbol    \beginargs{#1}\allargs}
\mathcommand\Animalppp      [1]{\Animalpsymbol  \beginargs{#1}\allargs}
\mathcommand\Maleppp        [1]{\Malepsymbol    \beginargs{#1}\allargs}
\mathcommand\Boyppp         [1]{\Boypsymbol     \beginargs{#1}\allargs}
\mathcommand\Girlppp        [1]{\Girlpsymbol    \beginargs{#1}\allargs}
\mathcommand\Femaleppp      [1]{\Femalepsymbol  \beginargs{#1}\allargs}
\mathcommand\Roundppp       [1]{\Roundpsymbol   \beginargs{#1}\allargs}
\mathcommand\Bishoppp       [1]{\Bishopsymbol   \beginargs{#1}\allargs}
\mathcommand\Quadrangularppp[1]{\Quadrangularpsymbol  \beginargs{#1}\allargs}
\mathcommand\Metppp[2]{\Metpsymbol     \beginargs{#1}\separgs{#2}\allargs}

\newcommand\bound     {{\rm bound}}
\newcommand\free      {{\rm free}}

\mathcommand\Vtripleindex[3]{\V\!_{{#1},\,{#2},\,{#3}}}
\mathcommand\Vdoubleindex[2]{\V\!_{{#1},\,{#2}}}
\mathcommand\Vsingleindex[1]{\V\!_{{#1}}}

\mathcommand\Erel[1]{\Gammaoffont\!_{#1}}
\mathcommand\Urel[1]{\Deltaoffont_{#1}}

\newcommand\vc{vari\-able-con\-di\-tion}

\mathcommand\theRprimefromstrongtoweak{
  \inparenthesesinlinetight{
     \domres\id{\Vwall\cup\Vsome\setminus\RAN\varsigma}
     \nottight{\nottight\uplus}
     \reverserelation\varsigma
  }
  \nottight{\circ}
  \ranres
    {\transclosureinline R}
    {\Vwall\cup\Vsome\setminus\RAN\varsigma}
  \nottight{\nottight{\nottight{\uplus}}}
  \Vsome\tighttimes\Vsall
}

\mathcommand\deltaminus{\delta^-}
\mathcommand\deltaplus{\delta^+}
\mathcommand\deltaplusplus{\delta^{+^+}}
\mathcommand\deltastar{\delta^*}
\mathcommand\deltastarstar{\delta^{*^*}}

\mathcommand\Vall     {\Vsingleindex\indexdelta         }
\mathcommand\Vwall    {\Vsingleindex\indexdeltaminu     }
\mathcommand\Vsall    {\Vsingleindex\indexdeltaplus     }
\mathcommand\Vgsome   {\Vsingleindex\indexgammaplus     }
\mathcommand\Vsome    {\Vsingleindex\indexgamma         }
\mathcommand\Vfree    {\Vsingleindex\indexfree          }
\mathcommand\Vbound   {\Vsingleindex\indexbound         }
\mathcommand\Vsomesall{\Vsingleindex\indexgammadeltaplus}

\mathapplycommand\VARall      {\VARsingleindex\indexdelta         }
\mathapplycommand\VARwall     {\VARsingleindex\indexdeltaminu     }
\mathapplycommand\VARsall     {\VARsingleindex\indexdeltaplus     }
\mathapplycommand\VARgsome    {\VARsingleindex\indexgammaplus     }
\mathapplycommand\VARsome     {\VARsingleindex\indexgamma         }
\mathapplycommand\VARfree     {\VARsingleindex\indexfree          }
\mathapplycommand\VARbound    {\VARsingleindex\indexbound         }
\mathapplycommand\VARsomesall {\VARsingleindex\indexgammadeltaplus}
\mathcommand\displayVARsall[1]{\VARsingleindex\indexdeltaplus
\!\!\!\:\left(\begin{array}{@{}c@{}}#1\end{array}\right)}

\mathcommand\rigidvari     [2]{#1_{#2}^\indexgammadeltaplus}
\mathcommand\existsvari    [2]{#1_{#2}^\indexgamma    }
\mathcommand\forallvari    [2]{#1_{#2}^\indexdelta    }
\mathcommand\freevari      [2]{#1_{#2}^\indexfree     }
\mathcommand\wforallvari   [2]{#1_{#2}^\indexdeltaminu}
\mathcommand\sforallvari   [2]{#1_{#2}^\indexdeltaplus}
\mathcommand\gexistsvari   [2]{#1_{#2}^\indexgammaplus}
\mathcommand\boundvari     [2]{#1_{#2}}
\mathcommand\vari          [2]{#1_{#2}}
\mathcommand\wforallvarilow[2]{#1_{#2}^
{\raisebox{-.82ex}{\math\indexdeltaminu}}}

\newcommand\indexhelper[1]{{\scriptscriptstyle#1\:\!\!}}
\newcommand\indexdeltaplus
{\indexhelper{\delta^{\raisebox{-.17ex}{\fvesf\hskip-0.14em +}}}}
\newcommand\indexdeltaminu
{\indexhelper{\delta^{\mbox{\fvesf\hskip-0.14em\rule[.2ex]{.7em}{.15ex}}}}}
\newcommand\indexgammaplus
{\indexhelper{\gamma^{\mbox{\fvesf\hskip-0.14em +}}}}
\newcommand\indexgammadeltaplus
{\indexhelper{\gamma\delta^{\raisebox{-.17ex}{\fvesf\hskip-0.14em +}}}}

\newcommand\indexdelta{\indexhelper\delta}
\newcommand\indexgamma{\indexhelper\gamma}
\newcommand\indexfree
{{\scriptscriptstyle\free}}
\newcommand\indexbound
{{\scriptscriptstyle\bound}}

\newcommand\Wellfsymb{\ident{Wellf}}
\mathapplycommand\Wellfpp{\Wellfsymb}

\mathcommand\beginargs{(}
\mathcommand\allargs  {)}
\mathcommand\separgs  {,\,}
\mathcommand\tightsepargs{,}

\mathcommand\minusppnoparentheses  [2]{{#1}\,\minussymbol\,{#2}}
\mathcommand\tightminusppnoparentheses  [2]{{#1}\minussymbol{#2}}
\mathcommand\divideppnoparentheses [2]{{#1}\,\dividesymbol\,{#2}}
\mathcommand\plusppnoparentheses   [2]{{#1}\,\plussymbol \,{#2}}
\mathcommand\plusppnoparenthesesi  [3]{{#2}\,\plussymbol^{#1}\,{#3}}
\mathcommand\tightplusppnoparentheses   [2]{{#1}\plussymbol{#2}}
\mathcommand\timesppnoparentheses  [2]{{#1}\,\timessymbol\,{#2}}
\mathcommand\undppnoparentheses    [2]{{#1}\und            {#2}}
\mathcommand\oderppnoparentheses   [2]{{#1}\oder           {#2}}
\mathcommand\impliesppnoparentheses[2]{{#1}\implies        {#2}}
\mathcommand\leqinfixppnoparentheses[2]{{#1}\,\tight\leq\,{#2}}
\mathcommand\geqinfixppnoparentheses[2]{{#1}\,\tight\geq\,{#2}}
\mathcommand\dividepp [2]{(\divideppnoparentheses {#1}{#2})}
\mathcommand\minuspp  [2]{(\minusppnoparentheses  {#1}{#2})}
\mathcommand\pluspp   [2]{(\plusppnoparentheses   {#1}{#2})}
\mathcommand\timespp  [2]{(\timesppnoparentheses  {#1}{#2})}
\mathcommand\undpp    [2]{(\undppnoparentheses    {#1}{#2})}
\mathcommand\oderpp   [2]{(\oderppnoparentheses   {#1}{#2})}
\mathcommand\impliespp[2]{(\impliesppnoparentheses{#1}{#2})}

\setnumberbibstyle
\renewcommand\mybibbaselinestretch{1.0}
\setmybibitemsep{0.2\topsep}
\renewcommand\mybibsection[1]{\mysection #1}
\renewcommand\referencessize{\normalsize}
\catcode`\@=11
\input ENDNOTES.sty
\catcode`\@=12
\def\enotesize{\normalsize}
\def\endnoteendsep{\vspace{2.5\topsep}}
\let\footnote=\endnote
\renewcommand\namefont{\sc}
\date
{\SEKIedition
 {\small Submitted March\,21, 2006}\\
 {\small First Print Edition August\,25, 2006}\\
 {\small Very Minor Improvements \Jul\,16, 2008}\\
 {\small Very Minor Improvements \May\,11, 2010}\\
 {\small Minor Improvements \Aug\,31, 2010}\\%
}
\begin{document}\makecover\maketitle\begin{abstract}%
\noindent 
In this position paper we briefly review the development 
history of\emph{automated inductive theorem proving} and\emph
{computer-assisted mathematical induction}. 
We think that the 
current low expectations on progress in this field result
from a faulty narrow-scope historical projection. 
Our main motivation is to explain
---~on an abstract but hopefully sufficiently 
descriptive level~---
why 
we believe that future progress in the field is to result
from human-orientedness and\emph\descenteinfinie.
\end{abstract}
\vfill\pagebreak
\setcounter{tocdepth}{1}\tableofcontents
\vfill\pagebreak
\section{Introduction}\label
{section Introduction}

\subsection{Subject Area}
In this \daspaper\ we are concerned with\begin{itemize}\noitem\item{\em
automated inductive theorem proving}\/ and\noitem\item{\em
computer-assisted mathematical induction}.\noitem\end{itemize}
Both terms refer to the
task of doing mathematical induction with the computer. 
The former term puts emphasis on the importance of strong automation
support, as found in the classical systems \nqthmwithcitation, 
\inkawithcitation, and \acltwowithcitation\ based on explicit induction. 
The latter and more general term, however, 
is to denote 
more human-oriented approaches in addition,
as found in \quodlibetwithcitation\ and other future systems
based on\emph\descenteinfinie. 
Note that we do not believe in the usefulness of the extreme representatives
of any of the two terms: 
Neither mere black-box automation nor mere proof-checkers 
can be too useful in mathematical induction.
Above that, we think that a successful system has to put strong emphasis
on both aspects and find a way to be both human-\emph{and} machine-oriented.

\subsection{Expectations and Importance of Future Progress}
A majority of researchers in the area of
{computer-assisted mathematical induction} seem to believe that 
no further progress 
can be expected in this area within the nearer future.
Moreover, recently, 
between two talks at a conference, one of the leading German senior
researchers in the field told me that 
he thinks that currently
it is hardly possible to get any funding for research on
{computer-assisted mathematical induction}. 

Thus, we should ask for possible scientific reasons for the current 
funding situation. We ought to check the justification of the belief that 
progress in computer-assisted mathematical induction 
is unlikely to occur in the nearer future.

It is, however, obviously not the case that progress 
in {computer-assisted mathematical induction} is considered to be unimportant.
Indeed, progress in computer-assisted mathematical
induction is in high demand for \maslong s, for verification of
software and hardware, and for synthesis of recursive programs.
Due to a slow-down in progress of automated mathematical induction
in the last decade, however, currently there does not seem to be
 much hope among scientists for further progress in the nearer future.

\subsection{A Possible Way to Future Progress \ --- \ Overall Thesis}\label
{subsection A Possible Way to Future Progress --- Thesis}
To show a possible way to future progress is the aim of this position paper.
Namely, to explain why we are confident that\emph\descenteinfinie\ 
can initiate a further breakthrough in computer-assisted mathematical induction.

\begin{sloppypar}
Together with\emph\descenteinfinie\ we present our ideas on the importance
of\emph{human-oriented theorem proving}, a point of view we have been 
holding and furthering for more than a dozen years \nolinebreak\cite
{wirthdiss,wirthlecture}.

``Human-oriented theorem proving'' basically means
that
---~to overcome the current stagnation~---
we have to develop\emph
{paradigms and systems} for the synergetic combination and cooperation of 
the human mathematician with
its semantical strength and the machine with its computational strength.

Our thesis is that\emph\descenteinfinie\ is such a paradigm.
\end{sloppypar}

\subsection{Organization of this \DasPaper}
The \daspaper\ organizes as follows.
In \sectref{section requirements specification}
we describe the general context where and why mathematics 
and mathematicians should win from computers.\,
As the reason for the little hope in progress in mathematical induction
seems to be a wrong projection from the past into the future,\, 
we cannot reasonably state
what we may hope to achieve by\emph{human-orientedness} 
and\emph\descenteinfinie\ 
(\sectref{section What Can we Hope to Achieve}) \ 
and why the two belong together
(\sectref{section Why Mathematical Induction?}) \ 
before we have had a short look at the history of 
computer-assisted theorem proving in
\sectref{section Short History of Computer-Assisted Theorem Proving}\@. \ 
Without diving too deep into technical details,
after presenting\emph\descenteinfinie\ 
(\sectref{section descente infinie}) \ 
and\emph{explicit induction} 
(\sectrefs{section explicit induction}
 {section Why Sticking to Explicit Induction Blocks Progress}), \ 
we then support our overall thesis of 
\sectref{subsection A Possible Way to Future Progress --- Thesis}
in \sectrefs{section support}{section The Fundamental}, \ 
and discuss the standard objections in 
\nolinebreak\sectref{section Discussion}\@. \ 
Finally, we conclude in \nolinebreak\sectref{section Conclusion}.
\vfill\pagebreak

\section{Requirements Specification}\label
{section requirements specification}
From the ancient Greeks until today,
mathematical theories, notions, 
and proofs are not developed the way they are documented. 
This difference is not only due to the iterative deepening 
of the development and the omission of easily reconstructible parts.
Also the global order of presentation in publication 
more often than not differs from the order of development.
This results in the famous\emph{eureka} steps, which 
puzzle the freshmen in mathematics.
The difference does not only occur in scientific publications where 
the\emph{succinct presentation of results} may justify this difference, 
but also for the vast majority of 
textbooks and lectures where the objective should be \begin{itemize}\item
to\emph{teach how to find} proofs, notions, and theorems.\end{itemize}
The conventional natural-language representation 
of mathematical proofs in advanced theoretical journals
with its intentional vagueness \cite[\litsectref{6.2}]{nonpermut} 
and hidden sophistication
can only inform highly educated human beings
about already found proofs.
This conventional representation, however
---~as fascinating as it is as a
summit of the ability of the human race to communicate deep structural
knowledge effectively~---
does not tell much about 
the\begin{itemize}\item originally applied plans and methods 
of\emph{proof construction}\end{itemize} 
and does not admit
computers\begin{itemize}\item to {\em check for soundness}\/ and\item to
{\em take over the tedious, error-prone, computational, and boring parts}\/ 
of proofs.\end{itemize}
Obviously, a computer representation that admits the flexibility for and the 
support of the issues of all above items in parallel 
plus the computation of\begin
{itemize}\item different conventional\emph{natural language presentations} 
tailored to various purposes\end{itemize}is in great demand
and could increase the efficiency of working mathematicians tremendously.
\vfill\pagebreak

\section{Short History of Computer-Assisted Theorem Proving}\label
{section Short History of Computer-Assisted Theorem Proving}

\subsection{Formula Language and Calculi}\label
{subsection Formula Language}
Starting with the Cossists and \viete\ in the \nth{15} and \nth{16} centuries,
the formula language of mathematics and its semantics were adequately
and rigorously modeled by the end of the \nth{19}~century
in \peano's ideography \nolinebreak\cite{peanoiotabar}
and \frege's {\fraknomath\Begriffsschrift} \nolinebreak\cite{begriffsschrift}.

An adequate rigorous
representation that supports a working mathematician's\emph
{theorem proving}, however, has not been found until today.
But already now the formula language of mathematics and its semantics can 
provide a powerful
interface between human and machine.\footnote
{\label{footnote human and machine in quodlibet}For instance, 
 the basic paradigm of the
 human-oriented automated inductive theorem prover 
 \quodlibetwithcitation\ is the following:
 The working mathematician can feed the machine with his semantical knowledge 
 of the domain by stating lemmas, 
 and the machine can use these lemmas for sparse but
 deep proof search \nolinebreak\cite{sr200401,samoacalculemus,samoa-lemmas,samoa-phd}. 
 When this search fails, the graphical user interface presents
 a not too deep state of the proof where progress stopped
 to the mathematician in a carefully designed human-oriented calculus
 \nolinebreak\cite{quodlibet-cade,kuehlerdiss,wirthdiss}
 who may provide help with additional lemmas
 and other hints. It should be remarked, however, that the practical
 implementation of this paradigm is still more a task than an achievement.
 \Cfnlb\ \sectref{section The Fundamental} for more on this.}

The numerous logic calculi developed during the \nth{20}~century
were mostly designed to satisfy merely theoretical criteria,
but not to follow the theorem-proving procedures of working mathematicians.

{An important step toward human-oriented calculi was done 
by \gentzenname\ \gentzenlifetime\ 
when he used his structural insights to refine
his \ND\ calculi (which were close to natural-language mathematics) 
into sequent calculi \nolinebreak\cite{gentzen}. 
%
These calculi
meant a huge progress toward
an adequate human-oriented representation of a working mathematician's
deductive proof search.
Sequent and tableau calculi capture the reductive (analytic, top-down, backward)
reasoning from goals to subgoals directly in the essential calculus rules and 
the generative (synthetic, bottom-up, forward) reasoning from axioms to lemmas 
can be adequately realized with lemmatizing versions of the Cut rule 
\cite
{pds,quodlibet-cade,wirthdiss,wirthcardinal} 
(\cfnlb\ \noteref{note lemma application}). \ 
Based on \gentzen's sequent calculus
there has been further progress into this direction:
Free-variable calculi \cite{fitting,isabellehol,wirthcardinal}
admit to defer commitments
until the state of the proof attempt provides sufficient information 
for a successful choice.
Thereby they help the mathematician to follow his proof plans more closely
by overcoming premature witness decisions forced by \gentzen's original 
calculus.
Indexed formula trees \nolinebreak\cite{sergediss} admit the mathematician 
to focus immediately on the crucial proofs steps and defer the 
problems of \math\beta-sequencing and \math\gamma-multiplicity
\nolinebreak\cite{nonpermut}.}

\subsection{Automation}\label
{subsection Automation}
Starting in the 1950s,
there was great hope to automate theorem proving 
with the help of computers and machine-oriented logic calculi.
State-of-the-art fully-\underline automated \underline theorem \underline
provers of today (\tightemph{ATP}s, such as
\VAMPIRE\ \nolinebreak\cite{vampire} and \WALDMEISTER\ 
\nolinebreak\cite{waldmeister,loechnerdiss})
represent a summit in the history of creative engineering.
That ATP systems will never develop into
systems that can assist a mathematician in his daily work, however,
is a general consensus among their developers for more than a dozen years now.
The reason for this is the following:
\begin{quote}\em
The automatic theorem provers' search spaces are too huge for complete
automation and completely different 
from the search spaces of the working mathematicians,
who therefore can neither interact with these systems,
nor transfer their human skills to them.\end{quote}
Note that this does not mean that ATP systems are useless.
They already now provide a powerful basis for the automation
in \maslong s such as \omegawithcitation.\vfill\pagebreak

\subsection{Proof Planning}
At the end of the 1980s, the ideas to overcome the approaching dead end 
in ATP
were summarized under the keyword\emph{proof planning}. 
Beside its human-science aspects \nolinebreak\cite{science-of-reasoning},
the idea of {proof planning} 
\nolinebreak\cite{bundy-inductive-proof-planning,proofplanningsystems}
is to add smaller and more human-oriented\emph{higher-level search spaces}
to the theorem-proving systems
on top of the\emph{low level search spaces} of the logic calculi.
In the 1990s, the major proof-planning systems \oysterclamwithcitation,
\omegawithcitation, and \lambdaclamwithcitation\
seem to have been led astray by the hopes that with these additional levels
\begin{enumerate}\item the underlying logic calculus could be 
neglected,\footnote
{\begin{itemize}\item The\emph{\OYSTERCLAM\ system}
 \nolinebreak\oysterclamcitation\ 
 has to solve the very hard task of
 constructing proofs in the intuitionistic \loef\ type theory of \OYSTER,
 whereas the vast majority of mathematicians and ATP engineers
 would use transformations such as
 the one to the modal logic S4 
 \nolinebreak\cite{goedelcollected,fittingmodalandintuitionistic,wallen}
 to prove intuitionistic theorems.\item
 Proof planning  in the\emph{old\/ \OMEGA\ system} \nolinebreak\omegacitation\
 severely suffers from its
 commonplace natural deduction calculus,
 because it exports low-level tasks to higher levels of abstraction;
 these low-level tasks have turned out to be most problematic in practice
 because they can neither 
 be ignored nor properly treated 
 on the higher levels.
 \item
 The\emph{\LAMBDACLAM\ system} \nolinebreak\lambdaclamcitation\
 does not have any fixed logic level 
 at all.\end{itemize}} 
and,
\item 
instead of the working mathematician himself, it would be sufficient to
get his proof plans to the machine.\end{enumerate}

\subsection{Alternative Point of View in Proof Planning}
To the contrary of these hopes, we believe that progress in proof planning
and computer-assisted mathematical theorem proving
requires the  further development of
human-oriented state-of-the-art logic calculi,
which free the higher levels from unnecessary low-level commitments
and admit the mathematician to interact directly with the machine,
even when the automation of proofs fails on the lowest logic level.
\begin{quote}\em
We need both high-level top-down interactive 
proof development and bottom-up support from a state-of-the-art 
flexible human-oriented calculus with strong automation.
\end{quote}
The neglect of the logic calculus and human--machine interaction
is to be overcome in the system \ISAPLANNER\
\nolinebreak\cite{proofplanningsystems,isaplannertracing,isaplannerprototype}
and in the new \OMEGA\ system currently under construction 
\nolinebreak\cite{pds} 
by using the standard calculus of \ISABELLEHOL\ 
\nolinebreak\cite{isabelles-logics-hol,isabellehol}
and the new human-oriented calculus of \CORE\ \cite{sergediss}, respectively.

\subsection{Conclusion: Human-Oriented Automated Theorem Proving}
 The completely automatic generation of a non-trivial proof
 for a given input conjecture
 is typically not possible today
 and
 ---~contrary to the complete automation of chess playing~---
 will probably never be. 

Thus, beside some rare exceptions
 ---~as the automation of proof search 
     will always fail on the lowest logic level from time to time~---
 the only chance for automatic theorem proving to become
 useful for mathematicians is\emph{a synergetic interplay 
 between the mathematician and the machine}.

For this interplay,
it does not suffice to compute human-oriented representations
of machine-oriented proof attempts for interaction with a user interface
during the proof search.
Indeed, experience shows that
the syntactical problems have to be presented accurately
and in their exact form.
Thus
---~to \nolinebreak give the human user a chance to interact~---
the calculus\emph{itself} must be\emph{human-oriented}.

\section{What Can we Hope to Achieve? And How?}\label
{section What Can we Hope to Achieve}
After all that history of great original expectations and 
down-slowing progress, 
what can we reasonably hope for the nearer future?

As described in \sectref{subsection Formula Language}
and \noteref{footnote human and machine in quodlibet},
the formula language of mathematics and its semantics already now
provides a powerful
interface between human and machine.
But we still have to find 
a representation of mathematical proofs supporting the 
issues mentioned in \sectref{section requirements specification}, namely: 
machine assistance in and
teaching of proof search, proof planning, and theory development; 
automation of 
tedious, error-prone, computational, 
and boring parts of proofs and checking for 
soundness; and the computation of various natural language presentations.

As full automation cannot succeed within the current paradigm,
we have to follow the human mathematicians,
although we do not know much about their procedures and they 
hardly
know how to
explain them.\footnote
{{\bf(Teaching Proof Search Procedures in Mathematics Lectures)}\\
 In the best lecture course I ever attended, every lecture
 an emeritus professor came into the lecture hall and asked what he
 is expected to teach here. ``Analysis\,II!'' ``Do you know the theorem of
 so-and-so?'' ``What is that?'' ``\ldots'' ``No, we do not know that!''
 Then the emeritus gave a precise (but often incomplete) statement of the 
 theorem, discussed it, and (after the students had a clear idea on the 
 meaning of the theorem!) 
 started proof\emph{search}.
 In the lecture I learned most,
 he presented a proof that failed three times and was finally finished 
 successfully overtime, not before patching the theorem. 
 But this seems to be
 the best universities can give to their mathematics students today.
 (The missing systematics they had better learn from textbooks.)
 An apprentice is explained the easy procedures and shown the hard ones.
 Then, as we do not explain proof search to our students, 
 it is probably one of the hard ones. 
 Nevertheless, I do hope we will be able to do this some time.\vfill\pagebreak}

The first steps on this way are to give the mathematician the freedom to 
go his way and let the system assist him. Not the other way round as usual!
We are convinced of 
a potential success of 
the following development cycle:
\begin{itemize}\item
In a first step, informal and formal logical calculi and the user interfaces 
have to provide the freedom to use
all the required means in a human-oriented design,
and then,\item 
in a second step,
we have to learn the heuristics that admit a feasible proof search
from the mathematicians; by human learning in the beginning, hopefully by 
artificial-intelligence machine-learning later.
\end{itemize}
And the starting point ought to be a 
human-oriented, machine-oriented, flexible state-of-the-art calculus
\cite{sergediss,wirthcardinal} and an administration of 
proof tasks in a proof data structure \nolinebreak\cite{pds}.

\section{Why Mathematical Induction?}\label
{section Why Mathematical Induction?}
In this \sectref{section Why Mathematical Induction?}, 
we briefly explain why we see an affinity between
human-orientedness and mathematical induction 
and why this position paper is about both\emph\descenteinfinie\
and human-orientedness in parallel.

Beside some proof-theoretical peculiarities of 
mathematical induction that do not really have a
practical effect,\footnote
{{\bf(Proof-Theoretical Peculiarities of Mathematical Induction)}\par\noindent
The following often mentioned 
(\cf\ \eg\ \nolinebreak\cite[\litsectref 5]{bundy-survey})
proof-theoretical peculiarities of 
mathematical induction do not really have a
special practical effect on inductive theorem proving,
simply because efficiency problems cause the same effects already for the case of
deductive theorem proving:\begin{itemize}\noitem\item
As the theory of arithmetic is not enumerable (\cite{goedel,goedelcollected}),
completeness of a calculus
\wrt\ the standard notion of validity cannot be achieved.\par
In practice, however, it does not matter whether our proof attempt fails because
our theorem 
will not be enumerated ever or 
will not be enumerated before doomsday.
\item
By \nolinebreak\gentzen's Hauptsatz 
on Cut elimination \nolinebreak\cite{gentzen}
there is no need to invent new formulas in a proof of a
deductive theorem. 
Indeed, such a proof 
can be restricted to ``sub''-formulas of the theorem under consideration.
In contrast to lemma application (\ie\ Cut) in a deductive proof tree, 
the application of induction hypotheses and lemmas 
inside an inductive reasoning cycle cannot generally
be eliminated in the sense that the
sub-formula property could be obtained, \cf\nolinebreak\ \cite{induction-no-cut}.
Thus, for inductive theorem proving, ``creativity'' cannot be restricted to 
finding just the proper instances, 
but may require the invention of new lemmas and notions.
\par
Again, in practice, however, 
it does not matter whether we have to extend our proof search to 
additional lemmas and notions for 
principled reasons or for tractability \nolinebreak\cite{baazleitschcolllog}.
\end{itemize}}
mathematical induction is the area of mathematical theorem 
proving where our heuristic knowledge is best.
This is the case both for 
human (\tightemph\descenteinfinie, \cfnlb\ \sectref{section descente infinie}) 
and for machine-oriented heuristics
(\tightemph{explicit induction}, \cfnlb\ \sectref{section explicit induction}).
As these two heuristics are completely different in their surface structure
and the progress in practical usefulness was quite moderate in the last decade,
mathematical induction 
is a good area to look for evidence for our thesis on\emph
{human-orientedness}:\begin
{quote}\em\sloppy
Human-oriented procedures can overcome the current slowdown of progress
in computer-assisted theorem proving.
Their
---~even compared to machine-oriented procedures~---
huge search spaces can be controlled by heuristics
learned from human mathematicians working with advanced systems.\end{quote}
\vfill\pagebreak

\yestop\yestop\section{\em\DescenteInfinie}\label{section descente infinie}
In everyday mathematical practice of an advanced theoretical journal the 
frequent inductive arguments 
are hardly ever carried out explicitly.
Instead, the proof just reads something like 
``by structural induction on \math n, \qedabbrev'' or 
``by induction on \pair x y over \math <, \qedabbrev\closequotecomma
expecting that the mathematically educated reader could easily expand the 
proof if in \nolinebreak doubt. \ 
In \nolinebreak contrast, 
very difficult inductive arguments, sometimes covering several 
pages, such as the proofs of \hilbert's\emph
{first \math\varepsilon-theorem} \nolinebreak\cite[\Vol\,II]{grundlagen} \ 
or \gentzen's\emph{Hauptsatz} \nolinebreak\cite{gentzen}, \ 
or confluence theorems such as the ones in \nolinebreak\cite
{gwrta,wirthconfluence,wirth-shallow}  \ 
still require considerable ingenuity and\emph{will} be
carried out!
The experienced mathematician engineers his proof roughly according to
the following pattern:
\begin{quote}\howdescenteinfiniegoes\end{quote}\noindent\label{section items}%
The hard tasks of proof by mathematical induction are
\begin{description}\item[(Hypotheses Task) ]\mbox{}\\{to find the
 numerous induction hypotheses (as, \eg, 
 in the proof of \gentzen's Hauptsatz on Cut-elimination)}
 {and}
\item[(Induction-Ordering Task) ]\mbox{}\\{to construct 
 an\emph{induction ordering} for the proof, \ie\ 
 a \wellfounded\ ordering that satisfies
 the ordering constraints of all
 these induction hypotheses in parallel. (For instance, this was the hard
 part in the elimination of the \math\varepsilon-formulas
 in the proof of the \nth 1\,\math\varepsilon-theorem
 in \cite[\Vol\,II]{grundlagen}, \ 
 and in the proof of 
 the consistency of arithmetic by the \nlbmath\varepsilon-substitution
 method in 
 \nlbcite{ackermann-consistency-of-arithmetic}).}
\end{description}\par\noindent
The soundness of the above
method for engineering hard induction proofs 
is easily seen when the argument is structured as a proof by contradiction,
assuming a counterexample.
For \nolinebreak
\fermatname's \fermatlifetime\ historic reinvention of the method,
it is thus just natural that he developed 
the method itself in terms of assumed counterexamples \cite
{fermatslife,From-Fermat-to-Gauss,fermat-oeuvres,fermat-career,fermatsproof}. \ 
He \nolinebreak
called it\openquoteemph{\descenteinfinie\ ou ind\'efinie}\closequotefullstop
Here it is in modern language, very roughly speaking: \ 
A proposition \math\Gamma\ can be proved by\emph\descenteinfinie\ as follows:
\begin{quote}\em\howMethodofDescenteInfiniegoes\end{quote}\vfill\pagebreak
\par\noindent
There is historic evidence on\emph\descenteinfinie\ being the standard
induction method in mathematics:
The first known occurrence of\emph\descenteinfinie\ in history 
seems to be the proof of the irrationality of the golden number
\bigmaths{\frac 1 2\inpit{1\tight+\sqrt 5}}{} by the Pythagorean mathematician
\hippasosname\ (Italy) in the middle of the \nth 5~century 
\BC\ \nolinebreak\cite{hippa}.
Moreover, we find many occurrences of\emph\descenteinfinie\ in
the famous collection ``Elements'' of 
\euclidname\ \nolinebreak\cite{elements}\@.
The following eighteen centuries showed 
a comparatively low level of creativity in mathematical theorem proving,
but after \fermat's reinvention of the Method of\emph\DescenteInfinie\ 
in the middle of the \nth{17}~century, 
it \nolinebreak remained the standard induction
method of working mathematicians until today.

\yestop\noindent\thelogicalpointofview\par\yestop\noindent
For a more detailed 
discussion of\emph\descenteinfinie\ from the historical and linguistic
points of view
see \cite[\litsectref{2}]{fermatsproof}.\vfill\pagebreak

\section{Explicit Induction}\label
{section explicit induction}\label{sectionautomationeins}

In the 1970s, the\emph\Schoolofexplicitinduction\ 
was formed by computer scientists
working on the automation of inductive theorem proving.
Inspired by \robinsonname's resolution method \nolinebreak\cite{resolution}, 
they tried to solve problems of logical inference via 
reduction to machine-oriented inference systems.
Instead of implementing more advanced mathematical induction techniques,
they decided to 
restrict the \secondorder\ \theoremofnoetherianinduction\ \nlbmath{(\ident N)}
(\cfnlb\ \sectref{section descente infinie})
and the inductive Method of\emph\DescenteInfinie\ 
to \firstorder\emph{induction axioms} 
and deductive \firstorder\ reasoning
\nolinebreak\cite[\litsectref{1.1.3}]{wirthcardinal}\@. \ 

\yestop\noindent
Note that in these induction axioms,
the subformula
\par\noindent\LINEmaths{\theinnermostpartofW}{}\par\noindent
of \nlbmath{(\ident N)}
is replaced with a conjunction of 
instances of \nlbmath{\app P u}
with predecessors of \nlbmath v like in \nlbmath{(\ident S)}\@.
The induction axioms of explicit induction
must not contain the induction ordering \nlbmath <\@.

\yestop\noindent
Furthermore, note that although an induction axiom may take the form of a
\firstorder\ instance of the 
\secondorder\ \axiomofstructuralinduction\ \nlbmath{\inpit{\ident{S}}}
(\cfnlb\ \sectref{section descente infinie}), \ 
conceptually it is an instance of \nlbmath{(\ident N)}
and the whole concept of\emph{explicit induction} 
is a child of the computer,
whereas \inpit{\ident{S}} was already applied by the ancient Greeks 
\nolinebreak\cite{plato-induction}.

\yestop\noindent
The so-called ``waterfall''-method of the pioneers of this approach
\nolinebreak\cite{bm} refines this process into a fascinating heuristic,
and the powerful inductive theorem proving system \nqthmwithcitation\
has shown the success of this reduction approach already
in the 1970s. \ 
For comprehensive surveys on explicit induction 
\cfnlb\ \cite{waltherhandbook} and \cite{bundy-survey}\@. \ 
\Cfnlb\ \cite{wirthzombie} for a survey on the alternative approaches 
of implicit and inductionless induction.\footnote
{{\bf(Implicit and Inductionless Induction)}
\\Alternative approaches to automation of mathematical induction
 evolved from the\emph{\KNUTHBENDIX\ Completion Procedure} and were 
 summarized in the\emph\Schoolofimplicitinduction,
 which comprises 
 Proof by Consistency (Inductionless Induction),\emph\descenteinfinie\
 and implicit induction orderings (term orderings).
 Furthermore, there is pioneering work on the combination of induction
 and co-induction; \cf\ \eg\ \cite{padawitzexpanderzwo}.
 While Proof by Consistency and implicit induction
 orderings seem to be of merely
 theoretical interest today \nolinebreak\cite{wirthzombie},
 we \nolinebreak should carefully distinguish\emph\descenteinfinie\
 from the mainstream work on explicit induction.}

\yestop\noindent
\boyer\ \& \moore's \nqthmwithcitation\
and \bundy\ \& \hutter's rippling\footnote
{{\bf(The Idea of Rippling)}\par\noindent Roughly speaking,
the success in proving\emph{simple} theorems by induction automatically,
can be explained as follows:
If we look upon the task of proving a simple theorem 
as reducing it to a tautology,
then we have more heuristic guidance when we know that we probably 
have to do it by mathematical induction: Tautologies are to be found everywhere,
but the induction hypothesis we are going to apply can restrict the search
space tremendously. 
\par
In a famous cartoon of \bundyname's, the original theorem is symbolized as
a zigzagged mountainscape and the reduced theorem after the 
unfolding of recursive operators as a lake with ripples.
Instead of searching for an arbitrary tautology, we know that
we have to\emph{get rid of the ripples} to be able to apply an instance of
the theorem as induction hypothesis, as mirrored by the calm surface of the 
lake.\vfill\pagebreak} 
\nolinebreak\ripplingcitation\ are prime examples of 
practically useful automation-supported 
theorem proving and proof planning, respectively. \ 
Mainly associated with the development of explicit induction systems
such as \oysterclamwithcitation, \lambdaclamwithcitation, and
\inkawithcitation, 
there was still evidence for considerable improvements over
the years until the end of the 
\nth{20}~century \nolinebreak\cite{inductioncontest}. \ 
Since then, explicit induction has become a standard in education
in the \VERIFUN\ project 
 \nolinebreak\cite{verifun}. \ 
Today, the application-oriented explicit induction system
\acltwowithcitation\ is still undergoing some minor 
improvements. \ACLTWO\nolinebreak\
easily outperforms even a good mathematician
on the typical inductive proof tasks that arise 
in his daily work or as subtasks in software verification.
These methods and systems, however,
do not seem to scale up to hard mathematical problems 
and\emph{program synthesis} (where the computer-assisted 
inductive proof of a property of an underspecified program
actually is to synthesize the recursive definitions of the program). \
We believe that there are\emph{principled reasons} for this shortcoming.
\vfill\pagebreak

\section{Why Sticking to Explicit Induction Blocks Progress}\label
{section Why Sticking to Explicit Induction Blocks Progress}
\subsection{Flow of Information}
Apart from sociological reasons\commanospace\footnote
{{\bf(The Sociological Aspect of Explicit Induction as Normal Science)}
\par\noindent Another way in that explicit induction blocks scientific progress
 is a sociological one. 
 The heuristics to generate induction axioms in explicit induction
 have hardly changed since the end of the 1970s. 
 Some minor conceptual improvements 
 (such as \nolinebreak\cite{waltherLPAR92,waltherIJCAI93}, \eg)
 have turned out to be contra-productive in the practical context of a 
 highly optimized ``waterfall\closequotecomma
 because later phases were already optimized to patch the weaknesses of the 
 previous ones. 
 With all the men-power that went into explicit induction systems such as 
 \inkawithcitation\ or \acltwowithcitation,
 these systems have become so well-tuned to all simple standard 
 problems that it is hardly possible to demonstrate their
 shortcomings to referees within the time they are willing to 
 spend on the subject. 
 \par
 Beside that, to become competitive with \ACLTWO\ requires
 a common effort and years of work with little chance for 
 economic support or academic funding, approval, or rewards.
 In spite of this, mainly due to the idealism of 
 \kuehlername\ and \samoaname\ and a bunch of their students,\emph
 \descenteinfinie\ in \quodlibetwithcitation\
 ---~as explained in 
     \sectrefs{section support}{section Discussion}~---
 has already by now been able to outperform the formerly well-funded\emph
 {normal-science} \nolinebreak\cite{kuhn,wirth-kuhn} 
 \Schoolofexplicitinduction.} \hskip.15em
explicit induction blocks progress 
because it does not admit a {\em natural flow of information}\/ 
in the sense that a decision can be delayed or a commitment deferred
until the state of the proof attempt provides sufficient information 
for a successful choice. \
Indeed, \hskip.1em
explicit induction unfortunately must solve 
\thehardtasks\ \hskip.2em
already {\em before}\/ the proof has actually started. \
A proper induction axiom must be generated without
any information on the structural
difficulties that may arise in the proof later \nolinebreak on. \
For this reason, \hskip.1em
it is hard for an explicit-induction 
procedure to guess the right induction axioms for 
difficult proofs in advance. 

\subsection{Recursion Analysis and Induction Variables}
One of the most developed and fascinating applications of heuristic 
knowledge found in artificial intelligence, informatics, and computer science 
is\emph{recursion analysis} \nolinebreak\cite{bm}\@. \ 
This \nolinebreak is a technique for guessing a proper induction axiom 
by static analysis of the syntax of the conjecture and the recursive 
definitions. \ 
In this \daspaper, 
we subsume under the notion of ``classical recursion analysis''
also its minor improvements \nolinebreak\cite{stevens-rational-reconstruction,stevens-phd,walthertermination,waltherLPAR92,waltherIJCAI93}\@. \ 
Under the notion of ``recursion analysis'' 
we \nolinebreak also subsume\emph{ripple analysis}, an 
important extension of classical recursion analysis. \ 
\\\indent
{{Ripple analysis}\/
 is sketched already in \nolinebreak\cite[\litsectref 7]{bundy-recursion-analysis} and
 nicely described in \nolinebreak\cite[\litsectref{7.10}]{bundy-survey}\@. \ 
On the one hand, by rejecting
recursive definitions whose unfolding would block the application
of the induction hypothesis, ripple analysis
excludes some unpromising induction axioms of classical recursion analysis. \ 
On the other hand, by considering lemmas of a reductive character 
in addition to the actual recursive definitions, 
ripple analysis can find more useful
induction axioms than classical recursion analysis.} \ 
\\\indent
A \nolinebreak requirement, however, which we put on the notion of 
``recursion analysis'' is that it 
does not perform dynamical proof search but 
has a limited look{}ahead into the proof,
typically one rewrite step for each term in a set of subterms that
 covers all occurrences of\emph{induction variables}. \ 
Note that although ``induction variable'' is a technical term in
recursion analysis, roughly speaking, this notion is also common among
working mathematicians when they say that 
something is shown ``by
{induction on \nlbmath y}\,\closequotecomma for a variable \nlbmath y, 
for instance. \ 

\subsection{The Hypotheses Problem}
However fascinating and highly developed recursion analysis may get,
even the disciples of the \Schoolofexplicitinduction\ admit 
the inherent limitations of explicit induction: \ 
In \nolinebreak\cite[\p\,43]{protzenlazy}, 
we find not only small verification examples
already showing these limits, but also the conclusion:
\begin{problem}[{\cite[\p\,43]{protzenlazy}}] \ 
``We claim that computing the hypotheses {\em before} 
the proof is not a solution
to the problem and so the central idea for the lazy method 
is to postpone the 
generation of hypotheses until it is evident which hypotheses are required
for the proof.''\pagebreak\end{problem}
This ``lazy method'' removes only some limitations of explicit induction
as compared to\emph\descenteinfinie.
It focuses more on efficiency than on a clear separation of concepts,
and there is no implementation of it available anymore.
The labels ``lazy induction''
and ``lazy hypotheses generation'' that 
were coined in this context
are nothing but a reinvention of parts of \fermat's\emph\descenteinfinie\
by the explicit-induction community. 

\subsection{The Induction-Ordering Problems}
Computer scientists from the \Schoolofexplicitinduction\ used to consider
the tasks of\begin{itemize}\notop\item\headroom 
induction (\ie\ the choice of an induction axiom; \eg\ by recursion analysis) 
and\noitem\item deduction (\ie\ the rest of the proof; 
\eg\ by standard \firstorder\ reasoning techniques or by rippling 
\nolinebreak\ripplingcitation)\footroom\notop\end{itemize}to be orthogonal.
Working mathematicians know that this is wrong.
Especially the choice of a proper induction ordering interacts with the
several cases of a proof in such a way that a new proof idea tends to 
be in conflict with the induction ordering of the previous cases.
\begin{itemize}\item On the one hand, 
it is standard in explicit induction to fix induction orderings eagerly, at the 
very beginning of a proof.
\item On the other hand, 
fixing an induction ordering earlier than in the last steps 
of an induction proof has hardly any benefit ever:\begin{itemize}
\noitem\item
For\emph{difficult} proofs, this is obvious to any working mathematician.\item
For\emph{simple} proofs, the simple fact that any equation has a left- 
and a right-hand side provides us with sufficient pragmatics for searching
in that area of the search space where the smaller induction hypotheses
use to be applicable; provided that the specifier has written his specifications
in the standard style and the user has activated his lemmas for rewriting
with a suitable orientation.\end{itemize}\end{itemize}
\begin{problem} Explicit induction has to commit to a fixed and unchangeable
induction ordering eagerly, at the very beginning of an induction proof.
Such a commitment comes far too early and is a typical cause of failure.
Moreover, it is superfluous because there is 
hardly any heuristic benefit of committing 
to an induction ordering earlier than in the last steps 
of an induction proof.\end{problem}

\yestop\yestop\noindent
Beside the restriction of explicit induction to enforce an\emph{eager}
computation of induction axioms
(\ie\ induction hypotheses and the related induction orderings), 
explicit induction by recursion analysis has also another limitation:
\begin{problem}\label{problem no creative orderings}
Computing induction axioms
by recursion analysis can only result in
such induction orderings that are recombinations of orderings 
resulting from the recursive definitions 
(and from the currently known lemmas of a reductive character)
of the related specifications.
\end{problem}
As a matter of fact, most of the non-trivial induction proofs do not work out
with such induction orderings.
Moreover, for the case of program synthesis, 
we do not want to be restricted to such induction orderings.
For an instance of this see \nlbcite{bundy-gow},
where the quick-sort algorithm is to be synthesized 
from the requirements specification of the sorting function.
\vfill\pagebreak

\yestop\section{Why\emph\DescenteInfinie\ is Promising Now}\label
{section support}
\halftop\noindent The theoretical research paper 
\nolinebreak\cite{wirthcardinal} provides us with the {\em integration}\/
of {\em\descenteinfinie}\/ into deductive calculi. \
It is
---~to the best of our knowledge~---
the first such combination in the history of logic,
which 
does {\em not encode}\/ induction, but
actually {\em models}\/
the mathematical process of proof search by {\em\descenteinfinie\ itself}\/ 
and {\em directly}\/ supports it with the data structures required for a formal 
treatment.\footnote{%
 {(\bf On the Likeliness of Alternative Integrations of\emph
 \DescenteInfinie\ into State-of-the-Art Deductive Calculi)}\par\noindent
 I consider the integration of\emph\descenteinfinie\ into 
 state-of-the-art free-variable sequent and tableau calculi to be
 the my most important scientific contribution. \
 Since I actually have searched the whole 
 conceivable space of possible combinations
 far beyond what is documented in \nolinebreak\cite{wirthcardinal}, \hskip.2em
 I am pretty sure that my paper \nolinebreak\cite{wirthcardinal} 
 presents not only an
 elegant and both human- and machine-oriented
 combination of\emph\descenteinfinie\ and state-of-the-art deduction
 (including liberalized versions 
 (\deltaplus) of the \math\delta-rules), \hskip.2em
 but also the only possible one (up \nolinebreak to 
 isomorphism and beside some possible
 variations (\cfnlb\ \cite{wirthgreen}) and simplifications
 in formalizing \vc s) that actually {\em models}\/
 the mathematical process of proof search by {\em\descenteinfinie\ itself}\/ 
 and {\em directly}\/ supports it with the data structures required for a formal 
 treatment and does not {\em encode}\/ \encodingstext%
}

This integration 
(presented for state-of-the-art free-variable sequent and tableau calculi) 
is well-suited for an efficient interplay of human interaction and automation
and combines raising \nolinebreak\cite{miller}, 
explicit representation of dependence between free \math\gamma- and 
\math\delta-variables 
(according to \smullyan's classification \nolinebreak\cite{smullyan}),
the liberalized \mbox{\math\delta-rule,}
preservation of solutions, and unrestricted applicability
of lemmas and induction hypotheses. \
Moreover, the integration is natural in the sense that it goes together well 
with context-improved reasoning as in \nolinebreak\cite{sergediss},
with modern proof data structures as in \nolinebreak\cite{pds},
with program synthesis as in \nolinebreak\cite{bundy-gow},
and with logical binders such as \nlbmath\lambda\ and 
\nlbmath\varepsilon\ \nlbcite
{ahrendtgiese, grundlagen, leisenring, wirthhilbertepsilon,wirth-jal}. \ 
The semantical requirements for the integration are satisfied for 
practically all\footnote
{{\bf(Semantical Requirements of \cite{wirthcardinal})}\par\noindent 
 As described in \cite[\litsectref{2.1.4}]{wirthcardinal}
 all we need for the soundness of our integration of\emph\descenteinfinie\
 into two-valued logics are the validity of\begin{itemize}\noitem\item\sloppy
 the well-known\emph{Substitution \opt{Value} Lemma}
 (as, \eg, shown for different logics in \mbox{\cite[\litlemmref{3}]{andrewsA}},
  \mbox{\cite[\litlemmref{5401}(a)]{andrews}},
  \cite[\p\,127]{enderton},
  \cite[\p\,120]{fitting}, and
  \cite[\litpropref{2.31}]{fittinggod})
 and\noitem\item the trivial\emph{Explicitness Lemma} 
 (\ie\ the values of variables not explicitly 
 freely occurring in a term or formula have no effect on the value
 of the term or formula, \resp)
 (as, \nolinebreak\eg, 
  shown for different logics in \cite[\litlemmref{2}]{andrewsA},
  \cite[\litpropref{5400}]{andrews}, and
  \cite[\litpropref{2.30}]{fittinggod}).\vfill\pagebreak\noitem\end{itemize}} 
two-valued logics, such as clausal logic, 
classical first-order logic, and higher-order modal logic
\nolinebreak\cite[\litnoteref 8]{wirthcardinal}.

When computer-assisted inductive theorem proving started in the early 1970s,
the induction axioms of explicit induction were the only known
feasible formal means to integrate induction into deductive calculi.
Today, however, we are in a better situation 
because the results of \nolinebreak\cite{wirthcardinal} provide us 
with a simple, elegant, and both machine- and human-oriented integration 
of {\em\descenteinfinie\ itself}.

The only overhead this integration requires is to add a 
weight term to each sequent or proof\emph{goal}.
These weight terms stay inactive until a goal is applied as an induction
hypothesis. 
Compared to the application of a goal as lemma, 
such an induction-hypothesis application produces an additional ordering
subgoal, which asks us to show that the induction hypothesis is smaller 
than the goal to which it is applied 
in some \wellfounded\ ordering. 

\yestop\yestop\noindent
On a more technical level
---~to integrate\emph\descenteinfinie\ into a given logic calculus~---
we need
\begin{enumerate}\item to augment the goals (sequents) 
of the calculus with weight terms,
\item to add a lemma application\footnote
{\label{note lemma application}{\bf(Lemma Application)}\par\noindent
 {\em Lemma application}\/ works as follows.
Suppose that our proof goals consist of\emph{sequents}
 which are just disjunctive lists of formulas.
 (This is the simplest form of a sequent 
  that will do for two-valued logics.) \ 
When a lemma \bigmath{A_1,\ldots,A_m} 
is a subsequent of a sequent \nlbmath\Gamma\ to be proved
(\ie\ if, for all \math{i\in\{1,\ldots,m\}}, 
 the formula \nlbmath{A_i} is listed in \nlbmath\Gamma), 
its application 
closes the branch of this sequent ({\em subsumption}). \ 
Otherwise, the conjugates of the missing formulas \nlbmath{C_i}
are added to the child sequents (premises), one child per missing formula. 
This can be seen as Cuts on \nlbmath{C_i} plus subsumption.
More precisely
---~modulo associativity, commutativity, and idempotency~---
a sequent 
\bigmath{A_1,\ldots,A_m,B_1,\ldots,B_n}
can be reduced by application of the lemma
\bigmath{A_1,\ldots,A_m,C_1,\ldots,C_p} to the sequents 
\par\halftop\noindent\LINEnomath{
\bigmaths{\overline{C_1},A_1,\ldots,A_m,B_1,\ldots,B_n}{}
\hfill\math\cdots\hfill\bigmaths{\overline{C_p},A_1,\ldots,A_m,B_1,\ldots,B_n}.
}\par\halftop\noindent
In addition, roughly speaking,
any time we apply a lemma, we can instantiate its free variables
locally and arbitrarily.
\Cf\ \cite{wirthcardinal,nonpermut} for more on this.}
to the calculus if not already present,
\item to patch the lemma application of the calculus to admit
   induction-hypothesis application, which generates an additional 
   ordering subgoal for soundness based on the weights of 
   induction hypothesis (lemma) and goal, and
\item  to solve the ordering constraints of the induction-hypothesis
applications.\end{enumerate}
Typically, these requirements are easily satisfied,
although there may be problems with calculi based on fixed 
logical frameworks.\footnote
{{\bf
 (Integration of\emph\DescenteInfinie\ 
  into Logical Frameworks)}\par\noindent
 Item\,4 of the enumeration in \sectref{section support}
 is typically no problem 
 because we can get along with semantical orderings
 \nolinebreak\cite[\litdefiref{13.7}]{wirthdiss}\@. \ 
 Indeed, we do not need term orderings \nolinebreak\cite{simplificationorderings}
 anymore, contrary to what was the case with \QUODLIBET's predecessor 
 \UNICOM\ \nolinebreak\cite{unicom}\@. \ \par
 Items 1, 2, and 3, however, do not seem to be easily achievable
 with \ISABELLEHOL\ \nolinebreak\cite{isabelles-logics-hol,isabellehol},
 for instance\@. \ 
 A logical framework (such as \ISABELLE\ 
 \nolinebreak\cite{isabelle-generic,isabellesevenhundred,isabelle-reference}) 
 can hardly mirror 
 general mathematical activity,
 but only the logic calculi known at the time of its development. 
 This makes progress toward 
 human-oriented automatable calculi very difficult.
 As a convenient realization of\emph\descenteinfinie\ does not seem to be
 so easily possible in \ISABELLE-based systems,
 a lot of additional lemmas (or else ingenious recursive specification)
 may be necessary as described in \litsectref{1} (or else the solution) of 
 \nolinebreak\cite{nominal-techniques-isabelle-hol}\@. \ 
 Moreover, for the idea to support program synthesis via\emph\descenteinfinie\
 on the lower level of inductive theorem proving for software verification
 (\cfnlb\ our \sectref
  {section Why Sticking to Explicit Induction Blocks Progress} and 
  \nolinebreak\cite{bundy-gow}),
 the recursion facilities of \ISABELLEHOL\ seem to be insufficient:
 \slindname's 
 recursion theorems \nolinebreak\cite{slind-recursion}
 require termination proofs at a too early stage of 
 development \nolinebreak\cite{wirth-shallow}.}\vfill\pagebreak

\yestop\section
[The Fundamental Practical Advantage of  {\em\DescenteInfinie}]
{The Fundamental Practical Advantage of\\{\em\DescenteInfinie}}\label
{section The Fundamental}

\halftop\noindent
For human mathematicians, 
non-trivial mathematical proofs appear to have a {\em semantical}\/ nature. \
Therefore,
\maslong s
should comply with natural human proof techniques and 
should be able to follow the exact order in which the
human user organizes his semantical problem solving. 

Automated theorem proving, 
however, 
works on {\em syntactical}\/ domains. \
These syntactical domains are 
different from the semantical ones. \
Typically, they admit neither a global view on the proof task
nor the realization of a false commitment. \
For instance, if recursion analysis results in a useless induction axiom, 
the proof attempt fails completely.\footnote
{{\bf(Productive Use of Failure and Patching Faulty Conjectures)}\\
 Although, a failure of a proof is a complete one in case of
 a wrong induction axiom in explicit induction, 
 from such a failure, we might gain some insight 
 on the proof \nolinebreak\cite{failure-guide-induction} or 
 on the conjecture \nolinebreak\cite{protzendiss,protzenpatching}.
 And then we may start a more promising proof attempt 
 with different settings.\vfill\pagebreak} \
Of \nolinebreak course,
this does not mean that automatic search in a calculus is useless. \
To the contrary, 
an anytime and sparse automated syntactic search 
through the semantically highly 
redundant search spaces of a logic calculus
is most helpful in parallel to the interaction
of the human user. 

In such a parallel approach, 
the ``hot'' constraints should always be solved first. \
With the term \nolinebreak``{\em hot}\/ constraints\closequotecomma
we \nolinebreak mean constraints with solutions that are strongly indicated
by the current state of the proof attempt
in the sense that there is a committing
step toward their solution 
that makes a success of the proof attempt more likely
or without that the proof can hardly succeed. \
Although those constraints that are hot for a mechanic procedure
and those that are hot for a human mathematician 
in the construction of the proof idea will be different
more often than not, man and machine can cooperate very well,
provided that the constraints 
can be solved in any intended order and the effects 
can be communicated on the basis of a common view. \
Note that a step from either side will typically change the set
of hot constraints of the other.

We are very well aware of the fundamental difficulties and open questions
that have to be solved for such a cooperation of man and machine.
It actually cannot be denied that there seem to be several divergences 
between man and machine, especially:\begin{itemize}\item
Automation prefers fully expanded definitions while the human user prefers
a concise representation with composite notions.\item 
The higher the automatization the more difficult the analysis of a failed
proof attempt for the human user.\end{itemize}
Nevertheless, we are convinced that a cooperation of man and machine
on the basis of a common view is a realistic goal.

Now we finally just have to mention the fundamental practical advantage
of\emph\descenteinfinie\ as compared to 
encodings of \encodingstext:
\begin{quote}\em
The fundamental practical advantage of our integration 
of\emph\descenteinfinie\ is that the constraints of the 
inductive proof search can now be solved together with all other
constraints of the whole deduction in any suitable order.
\end{quote}
Thus, if recursion analysis shows us the proper way,
we can solve the constraints in the order according to
the heuristics of explicit induction.
But any other order is also possible. And we may delay
solving the harder constraints until the state of the 
proof attempt provides us with information sufficient for a successful choice.

\yestop\section{Discussion}\label{section Discussion}
\yestop\subsection{Paradigm Shift without Sacrifice \ --- \ Really?}
\yestop\noindent 
In blank opposition to our evaluation of\emph\descenteinfinie\ in
\sectref{section support} as promising, 
in the 1990s and still in the beginning of the \nth{21}~century,
some leading scientists from the explicit-induction community
used to claim
\begin{enumerate}\item[(1)]
that\emph\descenteinfinie\ would be too complicated to be
useful in practice, and \item[(2)] that
the proper induction axioms could be computed before the actual proof search
by a partial inspection of the proof in a specialized presentation 
different from the actual proof search with some advanced 
artificial-intelligence techniques 
\nolinebreak\cite{hutter-cade-nancy}.\end{enumerate}
\label{section quodlibet shot claim one}%
Claim\,(1) has already been falsified by the successful treatment of\emph
\descenteinfinie\ in the theorem prover \QUODLIBET\ \nolinebreak\cite{maslecture,quodlibet-cade,kuehlerdiss,kwspec,loechner-lpo,sr200401,samoacalculemus,samoa-lemmas,samoa-phd,wirthconfluence,wirthdiss,wirthcardinal,wirth-shallow}\@. \ 
Although \QUODLIBET\ does not use any induction axioms,
it is competitive with the
leading inductive theorem prover \acltwowithcitation,
with the practically important exception that \ACLTWO\ is so
efficiently implemented that it can be used for both 
verification\emph{and testing} of software.

\yestop\noindent
We believe that also Claim\,(2) is wrong and that we need the freedom 
to solve \thehardtasks\ \hskip.2em
in small portions spread over the
whole search of the actual proof. \ 
This belief was also confessed to by others in 
\nolinebreak\cite[\litsectref{4.5}]{kraan}
and in \nolinebreak\cite[\litsectref{13.4}]{gow-phd}, \ 
and there is further 
recent evidence for this in \cite[\litsectref{8}]{samoa-phd}. \ 
Even if Claim\,(2) were right and the proposed procedure feasible,
it \nolinebreak would still be an uneconomic procedure
because there is no need to plan the induction axiom
with specialized tools based on a special additional representation
before searching for the actual proof.

\yestop\noindent
The deeper reasons behind the Claims (1) and (2)
seem to be conservatism and the fear 
that the heritage of the 
great heuristic contributions to inductive theorem proving 
developed within the paradigm of explicit induction could be lost.
Although such losses are typical for paradigm shifts \cite{kuhn,wirth-kuhn},
the fear seems to be completely unjustified in our case:
\begin{itemize}\item
Theoretically,\emph\descenteinfinie\ includes explicit induction.\item
Practically, 
\QUODLIBET\ has shown that 
in our framework of\emph\descenteinfinie,
the heuristic knowledge of\emph{recursion analysis} in 
the field of explicit induction is still applicable, indispensable, and 
at least as useful as before. \ 
We will explain this in 
\nolinebreak\sectref{section recursion analysis in descente infinie}. \ 
That also rippling probably stays as important as before is sketched
in \nolinebreak\sectref{subsection Rippling and Descente Infinie}.
\end{itemize}\vfill\pagebreak

\yestop\yestop\yestop\yestop\knuthquotation\germantextsix
{\schillername; Wallensteins Tod,\\2.\,Aufzug, 2.\,Auftritt; Wallenstein}

\subsection{Schism in Minds \vs\ Schism in Systems}\label
{subsection schism}
\yestop\noindent
Actually, the 
 schism between explicit induction on the one side
 and\emph\descenteinfinie\ on the other,
 never really existed in the minds of most of 
 the leading scientists of the field,
 especially not since the year 1996 
 \nolinebreak\cite[\litsectref{4.2}]{wirthzombie}. \
 This expertise, however, has neither been 
 published nor communicated to the outside of the inner circle. \
 Moreover
 ---~contrary to the flexibility of the minds~---
 in the powerful inductive theorem prover \acltwowithcitation\
 and most other such systems
 this schism is still manifest:

\yestop\begin{problem}[No Natural Flow of Information in \ACLTWO]\label
{problem ACL2}\\
 The only way to get \ACLTWO\ to use an induction
 ordering which is not of the kind of \probref{problem no creative orderings}
 is to add a recursive function \nlbmath f
 terminating over this ordering and to hint
 the prover to use the ordering of its termination proof
 for the eager generation of a\emph{eureka} induction axiom.
 Note that the function \nlbmath f is typically nonsense and will 
 be used nowhere and especially not in the theorem, so that the 
 hint to use it is really necessary.
\end{problem}

\yestop\yestop
\subsection{The \Role\ of Recursion Analysis in\emph\DescenteInfinie}\label
{section recursion analysis in descente infinie}
\yestop\noindent
The cases where eager induction-hypotheses generation 
is needed to guide the proof into the right direction
(\cf\ \eg\ \cite[\litsectref{3.3}]{wirthcardinal}) are so rare in practice
that the current standard induction heuristic of the\emph\descenteinfinie\
system \QUODLIBET\ 
\nolinebreak\cite{quodlibet-cade,sr200401} generates 
induction hypotheses only lazily, whereas the case splits for the induction 
variables are done eagerly right at the beginning (after simplification).
The possibility to be lazy even simplifies recursion analysis
when different induction schemes are in conflict
because we do not have to merge them completely:
Compare \cite[\litsectref{8.3}]{kuehlerdiss} with  
the complicated problems of 
\nolinebreak\cite{waltherLPAR92,waltherIJCAI93}!

\yestop\noindent
Nevertheless, 
recursion analysis plays an important \role\ also in \QUODLIBET\ and 
in\emph\descenteinfinie\ in general.
Even without generating induction hypotheses 
and the induction ordering eagerly,
the case analysis suggested by recursion analysis
is of great heuristic value. Indeed, nothing is more helpful
than to know how to start the proof of a conjecture (after simplification).

\yestop\noindent
The recursion analysis 
in\emph\descenteinfinie\ is most useful for solving the following 
task of case analysis:\vfill\pagebreak
\begin{description}\item[(Task of Case Analysis on Induction Variables)]\mbox
{}\par\noindent
Which outermost universal 
variables of the (simplified) conjecture to are to be used as induction 
variables,
and which lemmas are to be used for the case analysis on the 
structure of the induction variables? 

For instance, 
which lemmas of the following form are to be chosen for our 
induction variables \nlbmath{\hastype m{\nat}}
and \nlbmath{\hastype l{\app\lists\nat}}
for a natural number and a list of natural numbers, respectively?
\par\math{m\tightequal\zeropp\nottight{\nottight\oder}
\exists\hastype n{\nat       }\stopq\inpit{
m\tightequal\spp n
}}\par\noindent
\par\math{
m\tightequal\zeropp\nottight{\nottight\oder}
\exists\hastype p{\app\lists\nat}\stopq
\inparenthesesoplist{
m\tightequal\prod p
\oplistund
\Everypp\Primesymbol p
}}\par\noindent
\par\math{l\tightequal\nilpp\nottight{\nottight\oder}
\exists\hastype n{\nat       }\stopq
\exists\hastype k{\app\lists\nat}\stopq\inpit{
l\tightequal\cnspp n k
}}\par\noindent
\par\math{l\tightequal\nilpp\nottight{\nottight\oder}
\exists\hastype n{\nat       }\stopq
\exists\hastype k{\app\lists\nat}\stopq\inpit{
l\tightequal\appendpp k{\cnspp n\nilpp}
}}\par\noindent
\end{description}

\yestop\noindent
Note that this task
is most critical for explicit induction,
because the eager induction-hypotheses generation 
fixes the result of this case analysis and makes
a later adjustment impossible.
In\emph\descenteinfinie, however, this task is 
non-critical because it serves only as a heuristic hint
on how to start proof search.
This is shown in the following example.

\yestop\yestop\begin{example}\label{example recursion analysis not critical}
\par\noindent
Consider the toy example of the even-predicate on natural numbers in the
clause
\par\noindent\LINEmaths{
    \Evenpp{\plusppnoparentheses x z}\comma
\neg\Evenpp{\plusppnoparentheses x y}\comma
\neg\Evenpp{\plusppnoparentheses y z}
}.\par\noindent
When recursion analysis based on \bigmaths{\plusppnoparentheses{\spp{v}}{w}
=\spp{\plusppnoparentheses{v}w}}{}
suggests a base case of 
\bigmath{x\boldequal\zeropp} and a step case of \bigmaths{x\boldequal\spp{x'}},
then a proof attempt by explicit induction fails.
\par\noindent
A proof attempt by\emph\descenteinfinie,
however, 
can go on with
a second case distinction on 
\bigmath{x'\boldequal\zeropp} and \bigmath{x'\boldequal\spp{x''}}
and actually proceed by the two base cases of \bigmath{x\boldequal\zeropp}
and \bigmath{x\boldequal\spp\zeropp}
and a step case of \bigmaths{x\boldequal\spp{\spp{x''}}}.
This, however, is not possible for explicit induction
based on any form of recursion analysis. \ 
Note that the three cases of 
\bigmaths{y\boldequal\zeropp},
\bigmaths{z\boldequal\zeropp},
and \bigmaths{y\boldequal\spp{y'}\und z\boldequal\spp{z'}}{}
provide yet another way for\emph\descenteinfinie\ to extend the proof attempt
into a successful proof.\end{example}
\vfill\pagebreak

\yestop\yestop\subsection{Rippling and\emph\DescenteInfinie}\label
{subsection Rippling and Descente Infinie}
\yestop\noindent
Although \QUODLIBET\ does not implement\emph{rippling} 
\nolinebreak\ripplingcitation\ yet
(but applies a less syntactically restricted search by a refined
contextual rewriting with markings 
\nolinebreak\cite{samoa-lemmas,samoa-phd}),
we expect that the restrictions
of the search space introduced by rippling can be more useful in the 
less restrictive framework of\emph\descenteinfinie\ than in the more
restrictive framework of explicit induction.

\yestop\yestop\noindent
When induction hypotheses are not generated eagerly,
``creational rippling'' \nolinebreak\cite{rippling} or 
``blowing up of terms'' \nolinebreak\cite{hutter-cade-nancy,coloringterms} 
are not required. Instead, the induction variables 
occur as additional sinks in the induction conclusion.\begin{itemize}\item
On the one hand, this makes rippling technically and intuitively simpler 
(\esp\ for destructor style recursion) and better suited for human--computer
interaction.\item
On the other hand, however, 
the induction variables in the conclusion must be somehow limited
in their character of being a sink: 
Unless we limit these sinks to swallow wave fronts consisting of destructors, 
we will have difficulties in finding
a \wellfounded\ induction ordering justifying the induction-hypothesis 
applications.
\end{itemize}
\yestop\yestop\subsection{Further Historical Limitations in Explicit Induction}
\yestop\noindent
Beside overcoming the must of generating induction axioms,
it should be noted that \QUODLIBET\ has some additional advantages
over classical explicit-induction systems:
\begin{itemize}
\yestop\item
The strong\emph{admissibility restrictions} of explicit induction
systems (\ie\ specification only by functional programs,
requiring their\emph{completeness and termination proofs in advance})
have shown to superfluous
\nolinebreak\admissibilitycitation. \ 
For the successful automation of inductive theorem proving
we do not have to enforce a complete specification 
of the the recursive functions that participate in the recursion analysis. \
Indeed, \QUODLIBET\ 
requires {\em neither termination nor completeness}\/ for
those recursive function definitions. \
Nevertheless, in \QUODLIBET, these recursive function definitions 
come with a guarantee on consistency and
are used for recursion analysis and other special heuristics. \
Thus, overspecification can 
be avoided and stepwise refinement of specifications becomes possible,
with a guarantee on the monotonicity of validity \nolinebreak\cite{wgcade}.

\yestop\noindent
This is of practical relevance in applications. \ 
For instance, in \cite{loechner-lpo}, \loechnername\
(who is not a developer but a user of \QUODLIBET) writes:
\yestop\begin{quote}``The translation into the input language of the inductive
theorem prover \QUODLIBET\ was straightforward. We later realized that 
this is difficult or impossible with several other inductive provers as
these have problems with mutual recursive functions and partiality'' 
\ldots\vfill\pagebreak\end{quote}
\yestop\item\sloppy
Another advantage compared to \ACLTWO\ with its poor user interface
and its restriction to a complete reset after failure
is the following: When automation fails, \QUODLIBET\ typically stops
early and presents the state of the proof attempt in a human-oriented form,
whereas everything is lost 
(and only some of the developers may know what 
 \nolinebreak to \nolinebreak do)
when explicit induction generates a useless
induction axiom (\cf\ \probref{problem ACL2} in 
\nolinebreak\sectref{subsection schism}).\end{itemize}

\yestop\yestop\subsection{Conclusion}
\yestop\noindent
Those researchers of the explicit induction community who realized what a
strong restriction it is to fix the induction axiom before the actual induction
proofs starts
---~the most important being
    \nolinebreak\cite{protzenlazy,protzendiss}, 
    \cite{gow-phd}, and \cite{bundy-gow}~---
always suffered from the 
wish to synthesize induction axioms. 
The same holds for the synthesis of simple recursive programs from their
inductive soundness proofs \nolinebreak\cite{hutter-cade-nancy,kraan} and the 
more general task of instantiating meta-variables of the input theorem,
where they also make sense as placeholders
for concrete bounds and side conditions 
of the theorem that only a proof can tell.
Indeed, the force to commit to a fixed induction axiom  eagerly 
is only acceptable for simple proofs or 
simple theorems without meta-variables.

\yestop\noindent
All in all, we have listed powerful arguments in 
\sectrefs{section support}{section The Fundamental}
and rebutted perceivable counter-arguments in this 
\sectref{section Discussion}. \ 
\vfill\pagebreak

\section{Conclusion}\label
{section Conclusion}
\yestop\subsection{Human-Orientedness}
\begin{sloppypar}
\halftop\noindent
As explained in \sectref{subsection Automation},
completely automated black-box theorem proving is approaching its
conceptual limits. \ 
Significant future progress will require a paradigm different from the 
artificial-intelligence
exploration of the huge search 
spaces of machine-oriented misanthropic calculi. \
Human-oriented theorem proving and human-oriented calculi 
provide the only known alternative and 
have been gaining more and more acceptance within the last dozen years. \
The major tasks in the intended advanced form of human--computer interaction
are
\begin{itemize}\item the further development of interface notions
following both hidden human cognitive concepts 
and the needs for powerful automation support, and\item the further improvement
of the exploitation of the semantical information 
for the syntactical search processes.
\end{itemize}
The basic paradigm of 
interaction should be an 
anytime search process that knows about the humans'
semantical strength and asks the human users for advice in their
area of competence before getting lost in complexity. \
With a human-oriented main-stream integration following this paradigm,
we can make man and machine a winning team.
\end{sloppypar}

\yestop\subsection{\em\DescenteInfinie}
\halftop\noindent
Induction axioms were 
never necessary for the working mathematicians and are not anymore necessary in
formalized mathematics or automated theorem proving
due to \nolinebreak\cite{wirthcardinal}. \
It now suffices to solve \thehardtasks\ \hskip.2em
in mathematics as well as in automated theorem proving.

There is no need to make the generation of induction axioms more flexible,
because we are in the lucky situation that we can 
have the cake {\em and}\/ eat it:
Indeed, 
we \nolinebreak can remove the restrictions induction axioms 
put on us and improve the usefulness of the
heuristic knowledge developed within the paradigm of explicit induction
at the same time.

When recursion analysis or eager induction-hypotheses generation 
show us the right way, we \nolinebreak can take it.
When they do not, we do not have to care for them.
We do not have to find a way to walk out of the maze of explicit induction.
We can fly over it.

After a proof has been completed, 
we can read out of it what the induction 
axioms would have been. 

As we do not need any induction axioms, however, 
we do not have to care at all whether our induction axioms should be\emph
{destructor style} or\emph{constructor style} or whatever mixed styles
one could imagine.

Moreover, note that
---~as discussed in \examref{example recursion analysis not critical}
    of \sectref{section recursion analysis in descente infinie}~---
the case analysis suggested by recursion analysis
is critical for the failure of explicit induction, 
but it serves only as a heuristic hint
on how to start proof search in\emph\descenteinfinie.

Beside the recursion analysis telling us how to start off
and beside the termination check of the induction ordering typically 
at 
the end,
we do not need any special procedures for induction. \
An induction-hypothesis application is just a lemma application 
generating an additional ordering subgoal.

{\em\Descenteinfinie}\/ and explicit induction
do not differ in the task 
(establishing inductive validity \nolinebreak\cite{wgcade})
but in the way the proof search is organized.
For simple proofs there is always a straightforward
translation between the two. 
The difference becomes obvious only for proofs of difficult theorems.

The results of 
\nolinebreak\cite{wirthcardinal}
on how to combine state-of-the-art deduction with\emph\descenteinfinie\
globally without induction axioms were not available when
explicit induction started in the early 1970s.
But now that we know how to do it, sticking to explicit induction as a must
is scientifically backward.
{\em\Descenteinfinie}\/ anyway admits a simulation of explicit induction that 
can profit from all the heuristics gathered in this field with the additional
advantages\begin{itemize}\item that
---~contrary to explicit induction 
   \nolinebreak\cite{bm,waltherLPAR92,waltherIJCAI93}~---
conflicting induction axioms do not have to be combined completely 
(because the major heuristic achievement of recursion analysis is 
 to tell which variables to start induction with, \cf\
 \examref{example recursion analysis not critical} 
 of \nolinebreak\sectref{section recursion analysis in descente infinie}), \ 
and\item that 
the induction ordering may stay open until the very end when all cases
of the proof are known (because an earlier fixing of the induction ordering
is hardly of any heuristic benefit ever).\end{itemize}
Both items are of great practical effect \nolinebreak\cite{sr200401,samoa-phd}.

\yestop\subsection{Summary}

\halftop\yestop\begin{quote}\em
While the heuristics developed within the paradigm of explicit induction remain the method of choice for routine tasks, explicit induction is an obstacle to progress in program synthesis and in the automation of difficult proofs, where the proper induction axioms cannot be completely guessed in advance. 
Shifting to the paradigm of \descenteinfinie\ overcomes this obstacle without sacrificing previous achievements.\yestop\end{quote}

\yestop\yestop\yestop\mysection{Acknowledgments}

\yestop\noindent
I would like to thank \huttername\ and 
\padawitzname\ for some very helpful advice,
and \samoaname\ for many fruitful discussions and a long list of most profound
critical suggestions for this \daspaper.
Moreover, it is a great pleasure to thank \tobias\ 
for the wonderful years we shared
working on \QUODLIBET\ and\emph\descenteinfinie.
\vfill\pagebreak

\yestop\yestop\yestop\yestop\yestop\yestop\yestop\mysection{Notes}\halftop
\halftop\yestop

\begingroup
\theendnotes
\endgroup

\yestop

\end{document}